\documentclass{article}

\usepackage{microtype}
\usepackage{graphicx}
\usepackage{subfigure}
\usepackage[T1]{fontenc}
\usepackage{libertine}

\usepackage{algorithm}
\usepackage{algorithmic}

\usepackage{hyperref}

\usepackage{amsmath}
\usepackage{amssymb}
\usepackage{mathtools}
\usepackage{amsthm}
\usepackage{natbib}
\usepackage{stmaryrd}
\usepackage[top=1in, bottom=1.5in, left=1in, right=1.2in]{geometry}
\usepackage{booktabs}

\makeatletter
\def\@fnsymbol#1{\ensuremath{\ifcase#1\or \or \or \or \or \or \fi}}
\makeatother

\usepackage[capitalize,noabbrev]{cleveref}

\theoremstyle{plain}
\newtheorem{theorem}{Theorem}[section]
\newtheorem{proposition}[theorem]{Proposition}

\theoremstyle{definition}

\theoremstyle{remark}

\newcommand{\opr}{\textsc{opr}}
\newcommand{\topr}{\textsc{topr}}
\newcommand{\ppo}{\textsc{ppo}}
\newcommand{\sft}{\textsc{sft}}
\newcommand{\tis}{\textsc{tis}}

\newcommand{\rmupos}{R_{\mu}^+}
\newcommand{\rmuneg}{R_{\mu}^-}

\newcommand{\murpos}{\mu_R^+}
\newcommand{\murneg}{\mu_R^-}

\newcommand{\cbar}{\, | \,}
\DeclareMathOperator*{\expect}{{\mathbb{E}}}
\newcommand{\kl}{\textrm{KL}}
\newcommand{\cdbar}{\, \| \,}
\newcommand{\bR}{\mathbb{R}}

\usepackage[textsize=tiny]{todonotes}

\begin{document}
\title{TAPERED OFF-POLICY REINFORCE\\ Stable and efficient reinforcement learning for LLMs}

\author{Nicolas Le Roux\thanks{\hspace*{-5mm}$*$ Equal contribution. \\$^\dagger$ Equal contribution. \\$^\nabla$ Corresponding author: nicolas@le-roux.name
\\$^3$ Work done at Reliant AI}$^{*, \nabla, 1}$\,\, Marc G. Bellemare$^{*,2}$\\Jonathan Lebensold$^{\dagger, 3}$\,\, Arnaud Bergeron$^{\dagger, 1}$\,\,Joshua Greaves$^{3}$ \\
Alex Fr\'echette$^{2}$\,\,
Carolyne Pelletier$^{2}$\,\,
Eric Thibodeau-Laufer$^{2}$\,\,\\
S\'andor Toth$^{2}$\,\,
Sam Work$^{2}$\,\,\\
$^1$Mila \,\,$^2$Reliant AI
}

\maketitle

\begin{abstract}
We propose a new algorithm for fine-tuning large language models using reinforcement learning. Tapered Off-Policy REINFORCE (TOPR) uses an asymmetric, tapered variant of importance sampling to speed up learning while maintaining stable learning dynamics, even without the use of KL regularization. TOPR can be applied in a fully offline fashion, allows the handling of positive and negative examples in a unified framework, and benefits from the implementational simplicity that is typical of Monte Carlo algorithms. We demonstrate the effectiveness of our approach with a series of experiments on the GSM8K and MATH reasoning benchmarks, finding performance gains for training both a model for solution generation and as a generative verifier. We show that properly leveraging positive and negative examples alike in the off-policy regime simultaneously increases test-time accuracy and training data efficiency, all the while avoiding the ``wasted inference'' that comes with discarding negative examples. We find that this advantage persists over multiple iterations of training and can be amplified by dataset curation techniques, enabling us to match 70B-parameter model performance with 8B language models.
As a corollary to this work, we find that REINFORCE's baseline parameter plays an important and unexpected role in defining dataset composition in the presence of negative examples, and is consequently critical in driving off-policy performance.
\end{abstract}

\section{Introduction}
Reinforcement learning (RL) and EM-type methods are rapidly becoming the dominant paradigm for fine-tuning large language models on complex tasks such as chain-of-thought reasoning. These methods can amplify a base model's performance without additional human data and can optimize for synthetic rewards \citep{zhang2024generative} and non-differentiable objectives \citep{chow2024inference}. While several popular methods rely solely on positive examples to fine-tune an LLM~\cite{zelikman2022star,gulcehre2023reinforced}, the ``trial and error'' nature of RL algorithms is especially well-positioned to leverage negative examples produced by the model, which are increasingly being recognized as key to efficient learning \citep{rafailov2023direct,tajwar2024preference,zhang2024negative,flet2024contrastive}. In fact, there is mounting evidence that the simplest of all methods, REINFORCE \citep{williams1992simple}, is a highly effective approach to fine-tuning LLMs \citep{ahmadian2024back}.

However, REINFORCE is essentially an on-policy algorithm. In the presence of negative rewards, its good behaviour can only be guaranteed when the training data distribution (or reference model) matches, or is close to the model's own distribution. This limits its ability to reuse past data, and puts pressure on the experimenter to select a just-right set of hyperparameters to avoid problems. Indeed, evidence of the instability of REINFORCE-type in off-policy training of LLMs can be found everywhere in the literature, from early work \citet{pang2021text} to key algorithmic choices made in the training of the Kimi~$\kappa$1.5~\citep{kimi25kimi} and DeepSeek-R1~\citep{deepseek} models. While KL regularization to the objective can mitigate this instability, it results in slower learning and requires additional hyperparameter tuning.

\begin{figure}
    \centering
    \begin{minipage}[c]{0.59\textwidth}
    \vspace{2em}
        \centering
        \includegraphics[width=\textwidth]{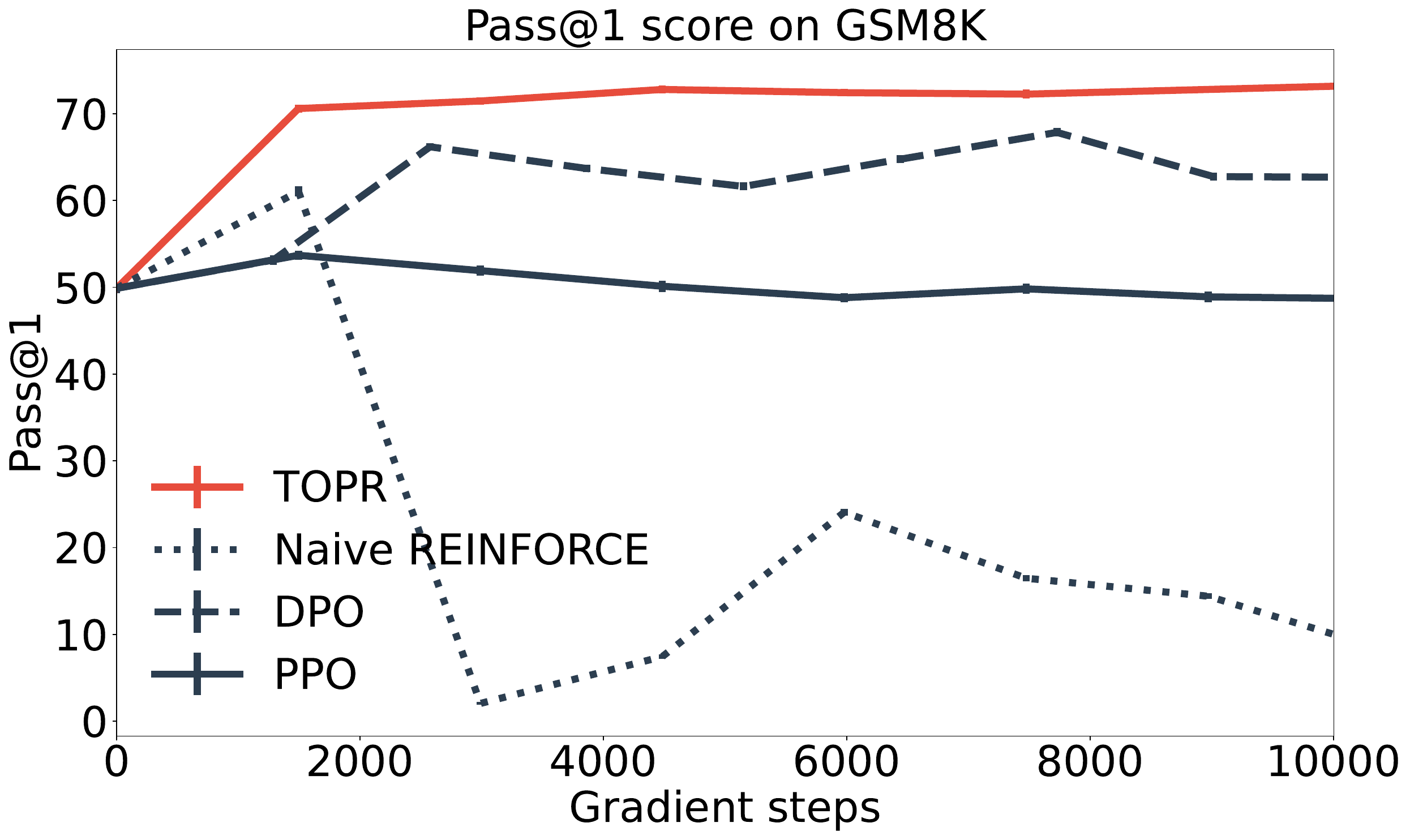}
    \hfill
    \vspace*{\fill}
\end{minipage}
\begin{minipage}[c]{0.39\textwidth}
\centering
\vspace*{\fill}
\begin{algorithm}[H]
\caption{TOPR (single iteration)}\label{alg:topr}
\begin{algorithmic}
\STATE \textbf{Input:} Language model $\pi$, reference $\mu$
\STATE \textbf{Input:} Prompts $x_1, \dots, x_m$
\STATE $y_i^1, \dots, y_i^n \sim \mu(\cdot \cbar x_i)$ for $i = 1, \dots, m$
\STATE Make dataset $\mathcal{D} = \{x_i, y_i^j \}$
\FOR{$(x,y) \in \mathcal{D}$}
        \STATE $\alpha = \left [ \tfrac{\pi(y \cbar x)}{\mu(y \cbar x)} \right ]_0^1$ if $R(x,y) < 0$ else $1$
        \STATE $\ell = \textsc{stop-grad}(\alpha) r(x,y) \log \pi(y \cbar x)$
        \STATE Perform gradient step w.r.t. loss $\ell$
\ENDFOR
\end{algorithmic}
\end{algorithm}
\vspace*{\fill}
\end{minipage}
        \caption{\textbf{Left: Test set accuracy (pass@1) on the GSM8K benchmark \citep{gsm8k} over the course of off-policy fine-tuning of the Llama 3 8B model.} As training becomes increasingly off-policy, the naive use of the REINFORCE gradient causes substantial performance degradation and PPO stops improving. DPO, which handles negative examples through a preference-based formulation, fares better but still falls well short of TOPR's performance. Pass@1 refers to the usual single-reasoning accuracy. 
        See Section~\ref{sec:results} for experimental details. \textbf{Right:} TOPR pseudo-code (Section~\ref{sec:topr}).
    }    \label{fig:gsm8k_G8b_SC_DPO+PPO+TOPR+Naive}
\end{figure}

As an alternative we propose \emph{Tapered Off-Policy REINFORCE} (TOPR), an algorithm that is stable even when the model differs substantially from the data distribution, while leveraging positive and negative examples to their fullest. TOPR improves a language model $\pi$ (or \emph{policy}) by means of the asymmetric policy gradient
\begin{align}
\label{eq:topr_grad}
    \nabla J_{\topr}(\pi) &= \sum_{\tau: R(\tau) \ge 0} \mu(\tau) R(\tau) \nabla \log \pi(\tau) + \sum_{\tau: R(\tau) < 0} \mu(\tau) 
 \left[\frac{\pi(\tau)}{\mu(\tau)}\right]_{0}^1 R(\tau) \nabla \log \pi(\tau) \; ,
\end{align}

where $\tau$ is a response (or \emph{trajectory}) sampled from some data-generating policy $\mu$, $R(\tau)$ is the reward associated with this trajectory, and
$[x]_a^b$ denotes the usual clipping function. Unlike other LLM algorithms in the REINFORCE family, TOPR does not require an explicit KL penalty to guarantee stable behaviour, making it both simpler to implement and computationally more efficient.
Compared to PPO \citep{schulman2017proximal}, DPO \citep{rafailov2023direct}, and the ``naive'' application of REINFORCE \citep{ahmadian2024back}, TOPR continues to improve reasoning performance even once $\pi$ differs substantially from $\mu$ (Figure~\ref{fig:gsm8k_G8b_SC_DPO+PPO+TOPR+Naive}). 

We characterize the stable off-policy performance of TOPR by using it to train language models to reason within the GSM8K and MATH benchmarks. We use these benchmarks to highlight the various factors at play in off-policy reinforcement learning of language models, in particular the importance of positive/negative balance in the training dataset. Surprisingly, we find that REINFORCE's baseline parameter -- commonly used as a variance reduction mechanism -- plays an altogether different role of balancing the dataset in this regime, and is essential to good performance. Critically, our results indicate that choosing the best baseline requires taking more than only the mean return into account. We conclude with a series of multi-iteration experiments showing the ability of TOPR to fine-tune language models well beyond their base benchmark performance.

\section{Off-policy policy optimization}
\label{sec:off_policy}

We consider an autoregressive language model $\pi$ that, given a prompt $x$, assigns a probability to a length-$n$ response $y$ according to
\begin{equation*}
    \pi(y \cbar x) = \prod_{i=1}^n \pi(y_i \cbar x, y_{<i}) .
\end{equation*}
Given a reward function $R(x,y)$ that measures the quality of the response $y$ to $x$ and a dataset of prompts $x_1, \dots, x_m$, we consider the problem of maximizing the expected reward
\begin{equation*}
    \label{eq:rl_supervised_loss}
    J(\pi) = \frac{1}{m}\sum_{j=1}^m \Big [ \expect_{y \sim \pi(\cdot \cbar x_j)} R(x_j,y) \Big ] .
\end{equation*}
In this paper we abstract the prompt-response relationship and view this problem through the lens of policy optimization, where $\tau$ is a trajectory produced by the language model (i.e., the policy). With mild abuse of notation, we thus write
\begin{equation*}
    J(\pi) = \expect_{\tau \sim \pi} R(\tau) .
\end{equation*}

The original REINFORCE algorithm~\citep{williams1992simple} maximizes $J(\pi)$ through the process of \emph{on-policy policy optimization}. In the simplest form of the algorithm, a single trajectory $\tau$ is sampled according to $\pi$, and the parametrized policy $\pi$ is updated according to the unbiased gradient estimate
\begin{equation}\label{eqn:reinforce_gradient_estimate}
    \nabla \hat J(\pi) = R(\tau) \nabla \log \pi(\tau),
\end{equation}
whose expectation is
\begin{equation*}
    \nabla J(\pi) = \expect_{\tau \sim \pi} R(\tau) \nabla \log \pi(\tau) .
\end{equation*}

In practice, training is rarely truly on-policy, for example because data is generated in a parallel, asynchronous fashion \citep{mnih16asynchronous} or in a separate ``sidecar'' process \citep{noukhovitch2024asynchronous,deepseek}. It is obviously also desirable to reuse trajectories throughout training, especially when generating these trajectories incurs a substantial computational cost or because they have been generated by a different process (e.g., expert trajectories). In the \emph{off-policy policy optimization} (OPPO) setting, we assume the existence of a reference distribution $\mu$, typically different from $\pi$, which produces training trajectories. Our main goal in this paper is to highlight the pitfalls of dealing with negatively-rewarded trajectories in off-policy policy optimization and propose a solution -- TOPR -- that avoids these pitfalls to produce performant, stable behaviour when training language models. By way of explaining the algorithmic choices behind TOPR, we review existing solutions and how they fall short of our desiderata.

\subsection{The problem with naive REINFORCE}\label{sec:negative_rewards}

As a warm-up, consider a binary reward function $R(\tau) \in \{-1, 1\}$ and the algorithm that samples a trajectory $\tau$ from $\mu$, then updates the policy $\pi$ according to Equation~\ref{eqn:reinforce_gradient_estimate}:
\begin{equation}\label{eqn:naive_reinforce_update_vanilla}
\nabla \hat J_{\mu}(\pi) = R(\tau) \nabla \log\pi(\tau) .
\end{equation}
This essentially corresponds to the ``naive'' off-policy application of the REINFORCE algorithm~\citep{ahmadian2024back}. In expectation, this update maximizes the objective
\begin{equation}\label{eqn:naive_reinforce_objective}
    J(\mu) = \sum_{\tau \in T^+} \mu(\tau) \log \pi(\tau) - \sum_{\tau \in T^-} \mu(\tau) \log \pi(\tau) \; ,
\end{equation}
where $T^+$ and $T^-$ are the set of trajectories with positive and negative rewards, respectively. 
The first term is maximized by making $\pi$ as close to $\mu$ as possible on the positive subset $T^+$. The second term, on the other hand, incentivizes $\pi$ being \emph{as far from $\mu$ as possible}.
This term is unbounded above (in terms of $\pi$) and can be made arbitrarily large by driving the probability of any single trajectory supported by $\mu$ to zero. This acts as a destructive force on the the model parameters, driving them to producing infinitely negative logits and without safeguards, eventually causes degenerate behaviour.\footnote{This issue doesn't appear in the on-policy application of REINFORCE because, by definition, a trajectory $\tau$ whose probability $\pi(\tau)$ is small is unlikely to arise in the dataset.}

We will show in Section~\ref{sec:analysis} that, while the issue can be mitigated by early stopping, the use of a baseline parameter, or KL regularization towards $\mu$, all of these modifications effectively work by fully or partially ignoring negative trajectories and thus limit the amount of learning that can be done off-policy.

\subsection{Supervised fine-tuning}
A simple solution to avoid the catastrophic failure of the model due to negative trajectories is to remove them from the dataset entirely. This can be interpreted as a form of reward-weighted supervised fine-tuning (SFT). The corresponding objective is
\begin{align*}
J_{\sft}(\pi) &= \sum_{\tau \in T^+} \mu(\tau) R(\tau) \log \pi(\tau) \; ,
\end{align*}
where the trajectory $\tau$ has weight $\mu(\tau) R(\tau)$ if $R(\tau)$ is positive, 0 otherwise. If we write $\murpos(\tau) \propto \mu(\tau) R(\tau)$, then $-J_{\sft}(\pi)$ is the cross-entropy loss between $\murpos$ and $\pi$.

Supervised fine-tuning in the usual sense \citep{ziegler2019fine} can be viewed as the special case where all positive rewards are equal to +1 and $\mu$ is fixed and independent of the language model.
More interesting is when the dataset is generated by the LLM itself, i.e.. $\mu$ is $\pi$ or close to $\pi$, or by another LLM, possibly with a filtering step to further enhance dataset quality. STaR \citep{zelikman2022star}, ReST \citep{gulcehre2023reinforced}, and ReST-EM~\citep{singh2023beyond} all follow this pattern.

By removing negative examples from the dataset, we obtain an objective that is bounded above. Consequently, these methods are stable. As they are implemented with a cross-entropy loss, they can also quickly learn to mimic the distribution $\murpos$, a characteristic that we retain in TOPR.
However, omitting negative examples clearly comes at an opportunity cost: for challenging problems, there may be few positive examples, and finding them may require additional machinery such as reference-guided grading \citep{zheng2023judging}, and wasted inference cycles. Mathematically, the lack of negative examples means that $\pi$ is incentivized to stay closer to $\mu$, limiting the amount of progress that can be achieved before having to resample from the LLM. 

\subsection{Truncated importance sampling}\label{sec:truncated_importance_sampling}

Importance sampling is perhaps the most common technique to address distribution shift. From
\begin{equation*}
    J(\pi) = \expect_{\tau \sim \mu} \Bigg [ \frac{\pi(\tau)}{\mu(\tau)} R(\tau) \Bigg ] \; 
\end{equation*}
we can derive an unbiased estimate of the on-policy gradient:
\begin{equation}\label{eqn:off_policy_reinforce_gradient}
    \nabla \hat J_{\opr}(\pi) = \frac{\pi(\tau)}{\mu(\tau)} R(\tau) \nabla \log \pi(\tau) .
\end{equation}
We call this the \emph{off-policy REINFORCE} (OPR) gradient. In theory, Equation~\ref{eqn:off_policy_reinforce_gradient} provides a convenient algorithm for optimizing the true objective $J(\pi)$: sample a trajectory $\tau \sim \mu$, and weight its update by the importance ratio $\tfrac{\pi(\tau)}{\mu(\tau)}$.
In practice, however, it is well-known that importance sampling is plagued with excessive variance. This especially problematic when optimizing over sequences, where the importance ratio is a product of many per-step ratios \citep{precup01offpolicy}. 
Gradient variance matters both for positive trajectories -- whose probability $\pi(\tau)$ increases during training -- but also negative trajectories, where a single excessive ratio can have a substantial destructive effect on the model parameters.

The variance issue can be mitigated by truncating the importance ratios, a technique already proven effective in value-based reinforcement learning \citep{munos16safe,espeholt2018impala} and REINFORCE applied to extractive question answering \citet{gao2022simulating}. The corresponding sample gradient is
\begin{align*}
    \nabla \hat J_{\tis}(\pi) &= \left[\frac{\pi(\tau)}{\mu(\tau)}\right]_0^1 R(\tau) \nabla \log \pi(\tau) \; ,
\end{align*}
where
\begin{equation*}
    \big [ x \big ]_{a}^{b} = \min(\max(x, a), b) .
\end{equation*}
Truncated importance sampling (TIS) is an integral part of TOPR. There are, however, situations where following the gradient of $J(\pi)$ is \emph{not desirable}, justifying further enhancements. To see this, note that when the importance ratio $\frac{\pi(\tau)}{\mu(\tau)}$ is close to 0, so is the norm of the gradient. Should this happen for trajectories with positive rewards, the model will take a long time to increase that trajectory's probability. This is not specific to importance sampling and is indeed an issue with the usual on-policy REINFORCE~\citep{kakade2001natural}.
We shall see in our experiments how TIS, while effective, is more sensitive to dataset composition and the choice of reward baseline and can lead to slower optimization.

\subsection{Other policy gradient methods}

Rather than attempt to ``fix'' the naive REINFORCE algorithm as in the previous sections, we may instead more directly change the objective to improve stability and performance. Before introducing TOPR, we discuss two popular methods, PPO and DPO, that take this approach.

\paragraph{PPO: truncating the objective.} PPO~\citep{schulman2017proximal}, one of the most widely used policy-based methods, optimizes the objective
\begin{align}\label{eqn:ppo_loss}
    J_{\ppo}(\pi) &= \expect_{\tau \sim \mu}\min\left(\frac{\pi(\tau)}{\mu(\tau)}R(\tau), \left[\frac{\pi(\tau)}{\mu(\tau)}\right]_{1-\epsilon}^{1+\epsilon} R(\tau)\right)
\end{align}
for $\epsilon \in (0, 1)$. This objective implements an asymmetric treatment of positive and negative rewards and is essentially composed of three parts (Fig.~\ref{fig:taper_function}). In the near on-policy setting, when only few updates are made before resampling trajectories, this can be quite effective; GRPO \citep{deepseek}, for example, modifies Equation~\ref{eqn:ppo_loss} with a batch-dependent baseline.

However, the PPO objective applies the importance ratio to the reward, rather the gradient; as a consequence, the gradient of $J_{\ppo}(\pi)$ becomes zero outside of the $[1-\epsilon,1+\epsilon]$ range, limiting its usefulness and potentially causing brittleness when more than a handful of updates are made before resampling trajectories. Similarly, the algorithm is not incentivized to reduce the relative probability of negative trajectories below $1-\epsilon$, limiting the potential improvement from $\mu$.
Although variants such as sPPO increase this robustness~\citep{vaswani2022general}, their performance still drops after a few tens or hundreds of updates.

\paragraph{DPO: balancing positives and negatives.} DPO~\citep{rafailov2023direct} works with pairs of trajectories and maximizes the weighted log probability ratio of these two trajectories. For positive and negative trajectories $\tau_w$ and $\tau_l$, respectively, the DPO objective is
\begin{equation*}
    J_{\textsc{dpo}}(\pi) = \log \sigma \left (\beta \log \frac{\pi(\tau_w)}{\mu(\tau_w)} -\beta \log \frac{\pi(\tau_l)}{\mu(\tau_l)} \right ) ,
\end{equation*}
where $\sigma$ is the sigmoid function. When rewards are either -1 or 1, DPO can be repurposed to handle negative and positive trajectories \citep{guo2024direct,calandriello2024human}, and is in fact well-suited to off-policy policy optimization \citep{noukhovitch2024asynchronous}.
However, DPO does not directly aim to maximize $J(\pi)$ and, with a finite number of trajectories, it is possible for the objective to increase while the probability of the positive trajectory decreases, as long as the probability of the negative trajectory decreases more. We shall see in our experiments that, while DPO indeed performs well off-policy, it is largely outperformed by TOPR.
More recently, CoPG~\citep{flet2024contrastive} also applied the idea of contrasting negatives and positives and, although their update shares a similar form with REINFORCE, the use of a carefully crafted baseline makes the method similar to DPO and IPO~\citep{azar2024general} and we therefore omit it from our analysis.

\begin{table}
    \centering
    \begin{tabular}{r||c|c|c|c||c|c|c|c}
    \multicolumn{1}{c}{} & \multicolumn{1}{c}{$a^+$} & \multicolumn{1}{c}{$b^+$} & \multicolumn{1}{c}{$a^-$} & \multicolumn{1}{c}{$b^-$} & \multicolumn{1}{c}{\parbox{15mm}{Negative examples}} & \multicolumn{1}{c}{\parbox{15mm}{Bounded objective}}& \multicolumn{1}{c}{\parbox{15mm}{Low\\variance}}& \multicolumn{1}{c}{\parbox{15mm}{Fast\\learning}} \\
    \toprule
    SFT                 &   1&  1&  0&  0&No&\textbf{Yes}&\textbf{Yes}&\textbf{Yes}\\
    Naive REINFORCE     &   1&  1&  1&  1 & \textbf{Yes} & No & \textbf{Yes}&\textbf{Yes}\\
    Off-policy REINFORCE                  &   0&  $+\infty$&    0   &$+\infty$ & \textbf{Yes} & \textbf{Yes}& No& No\\
    Truncated IS      &   0&  1&  0&  1   &\textbf{Yes}&\textbf{Yes}&\textbf{Yes}&No\\
    \midrule
    TOPR      &   1&  1&  0&  1   &\textbf{Yes}&\textbf{Yes}&\textbf{Yes}&\textbf{Yes} \\
    \bottomrule
    \end{tabular}
    \caption{TOPR combines the advantages of supervised fine-tuning, REINFORCE, and importance sampling to support stable and efficient off-policy fine-tuning of language models.}
    \label{tab:alg_comparison}
\end{table}

\section{TOPR: Tapered off-policy REINFORCE}\label{sec:topr}

We now introduce the TOPR algorithm. TOPR uses importance sampling to downweight negative trajectories that are not likely under $\pi$, while allowing positive trajectories to be upweighted irrespective of $\pi$. The general framework we consider involves two sets of truncation limits, $a^+ \le b^+$ and $a^- \le b^-$:

\begin{align}
\label{eq:topr_grad_general}
    \nabla J_{\cdot}(\pi) &= \sum_{\tau \in T^+} \mu(\tau)\left[\frac{\pi(\tau)}{\mu(\tau)}\right]_{a^+}^{b^+} R(\tau)\nabla \log \pi(\tau) + \sum_{\tau \in T^-} \mu(\tau)\left[\frac{\pi(\tau)}{\mu(\tau)}\right]_{a^-}^{b^-} R(\tau)\nabla \log \pi(\tau)  \; .
\end{align}
By choosing different truncation limits, we obtain many of the methods introduced in the previous section (Table~\ref{tab:alg_comparison}). 
TOPR itself corresponds to a range of truncation limits that combine the desirable properties of each of these methods into one learning rule. 

\paragraph{Gracefully unlearning negative trajectories.} By choosing $a^- = 0$, we allow the algorithm to progressively reduce the contribution of negative trajectories to the gradient, as provided by importance sampling. Any $a^- > 0$ must eventually lead to model degeneracy as with naive REINFORCE (Section~\ref{sec:negative_rewards}).

\paragraph{Quickly learning positive trajectories.} By choosing $a^+ > 0$, we gain the benefits of supervised fine-tuning: we ensure a minimum rate of learning for positive trajectories and accelerate their learning when they have a low probability under $\pi$. This allows us to avoid the ``quasi-local minima'' issue that plagues REINFORCE in high-dimensional action spaces. 

\paragraph{Trading off bias and variance.} The upper truncation limits allow us to keep gradient variance under control, as expected from truncated importance sampling. This is important early in training for negative examples, when a few examples may exhibit a very large importance ratio, and late in training for positive examples, where we expect the untruncated ratio to be much greater than 1.

\paragraph{}
We follow Occam's principle in defining the canonical form of TOPR as the algorithm where $a^- = 0$ and all other parameters are $1$. This yields the expected TOPR gradient:
\begin{align}
\label{eq:topr_grad_specific}
    \nabla J_{\topr}(\pi) &= \underbrace{\sum_{\tau \in T^+} \mu(\tau) R(\tau)\nabla \log \pi(\tau)}_{\textrm{SFT update for positive examples}} + \underbrace{\sum_{\tau \in T^-} \mu(\tau)\left[\frac{\pi(\tau)}{\mu(\tau)}\right]_{0}^{1} R(\tau)\nabla \log \pi(\tau)}_{\textrm{TIS update for negative examples}}  \; ,
\end{align}
which combines the SFT update for positive examples, leading to acceleration, and the TIS update for negative examples, allowing for their handling without brittleness. Algorithm~\ref{alg:topr} sketches out an implementation of TOPR in an off-policy, deep learning setting.

From a practical perspective, we will demonstrate in Section~\ref{sec:results} that this canonical parametrization is highly effective and provides robustness to the choice of data distribution and deep learning hyperparameters. Before doing so, however, we provide theoretical justification for the design choices behind TOPR.

\subsection{Analysis}\label{sec:analysis}

In Section~\ref{sec:negative_rewards} we argued that introducing a baseline parameter cannot create stable off-policy learning behaviour without risking sacrificing performance. We will make this point more precisely in this section and the next. To begin, let us revisit the expected naive REINFORCE update, now introducing a baseline parameter $c \in \bR$:
\begin{align}
\nabla J_{\mu, c}(\pi) &= \expect_{\tau \sim \mu} \Big [ \big (R(\tau) - c\big) \nabla \log\pi(\tau) \Big ] . \label{eqn:naive_reinforce_update}
\end{align}
We express this update in terms of the loss $\mathcal{L}_{\mu,c}$ that it implicitly minimizes:
\begin{align}
    \mathcal{L}_{\mu,c}(\pi) &= -J_{\mu,c}(\pi)\nonumber\\
    &= -J(\mu) - \expect_{\tau \sim \mu} \left [ (R(\tau) - c )\log \frac{\pi(\tau)}{\mu(\tau)} \right ] , \label{eqn:naive_reinforce_loss}
\end{align}
where we define $J_{\mu,c}$ so that $J_{\mu,c}(\mu) = J(\mu)$.
The following establishes the contribution of positive and negative examples and of the baseline to the expected loss $\mathcal{L}_{\mu,c}(\pi)$.
\begin{proposition}\label{prop:naive_reinforce_loss}
    Equation~\ref{eqn:naive_reinforce_loss} is the four-part loss
\begin{equation*}
    \label{eq:full_loss_with_kl}
    \mathcal{L}_{\mu,c}(\pi) = C + \rmupos \kl(\murpos \cdbar \pi) - \rmuneg \kl(\murneg \cdbar \pi) - c \kl(\mu \cdbar \pi) ,
\end{equation*}
    where $\murneg$ is the \emph{reward-weighted distribution}
    \begin{equation*}
    \murneg(\tau) = \left \{ \begin{array}{ll}
        \frac{\mu(\tau) |R(\tau)|}{\rmuneg} &         \text{if } R(\tau) < 0, \\
        0 & \text{otherwise;}
    \end{array} \right .
    \qquad 
        \rmuneg = \sum\limits_{\tau \in T^-} \mu(\tau) \big |R(\tau)\big|; \qquad T^- = \{ \tau : R(\tau) < 0 \},
\end{equation*}
and symmetrically for $\murpos$, and $C$ is a constant independent of $\pi$.
\end{proposition}
Proposition~\ref{prop:naive_reinforce_loss} shows that the baseline induces KL regularization towards ($c < 0$) or away $(c > 0)$ from the sampling distribution~\citep[see also][]{leroux2016efficient,vaswani2022general}. At $\mu = \pi$, this reaffirms the well-known fact that the baseline has no effect on the expected on-policy gradient \citep{williams1992simple,sutton99policy}.
In particular, when all rewards are positive ($\rmuneg$ = 0), Equation~\ref{eqn:naive_reinforce_update} moves the policy $\pi$ towards a reward-weighted version of the sampling distribution $\mu$ \citep{ghosh2020operator}. 

\begin{figure}
    \centering
    \includegraphics[width=\linewidth]
    {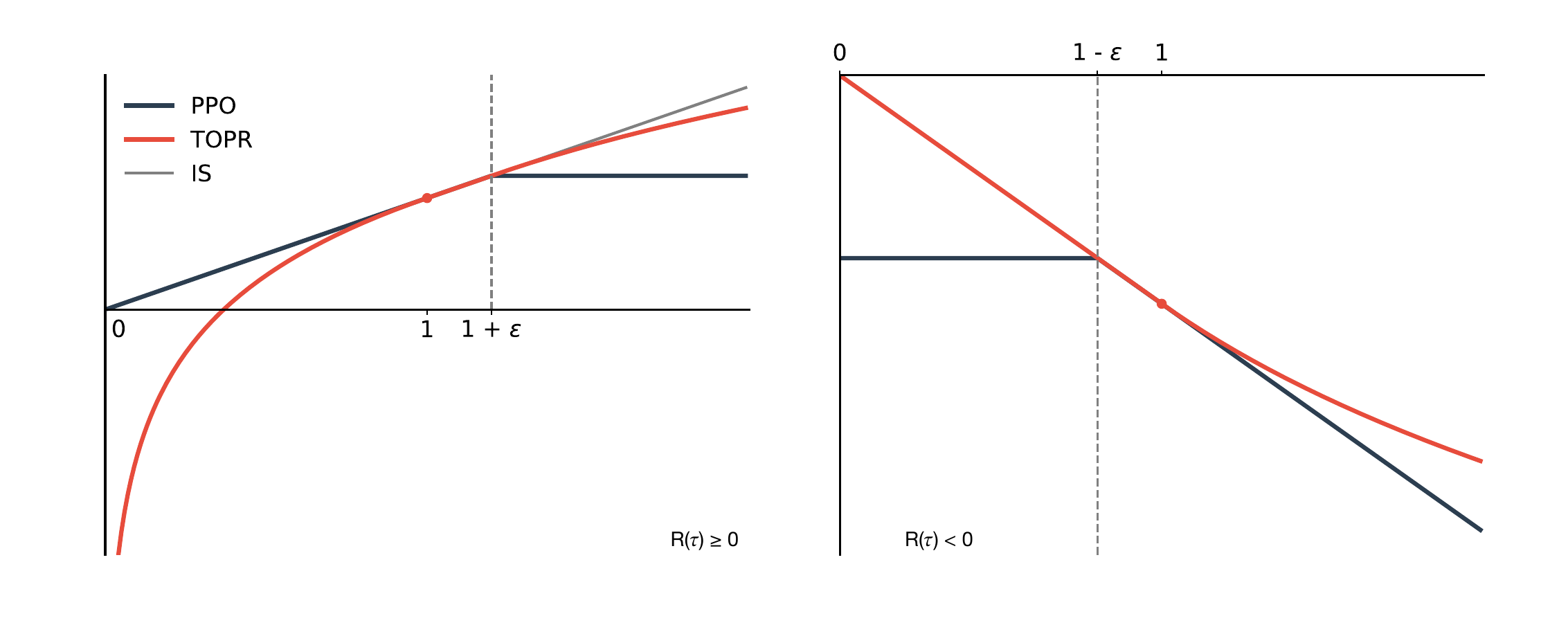}
    \caption{\textbf{Visualization the TOPR  objective (Equation~\ref{eqn:topr_loss} with $a^+ = 1, b^+ = 1 + \epsilon$, $a^- = 0, b^- = 1$) contrasted with PPO (Equation~\ref{eqn:ppo_loss}), as a function of the importance ratio $\tfrac{\pi}{\mu}$ (IS).} The two losses are equal on the intervals $[1, 1+\epsilon]$ and $[1-\epsilon, 1]$ (positive and negative rewards, respectively). However, PPO stops gradients when the ratio differs substantially from 1, which prevents it from being an effective off-policy algorithm. TOPR also implements a sharper positive-example loss for small importance ratios, accelerating the learning of these examples.
    \label{fig:taper_function}}
\end{figure}

Proposition~\ref{prop:naive_reinforce_loss} also shows that adding a baseline to minimize the impact of negative rewards $(c < 0)$ works by regularizing $\pi$ towards $\mu$. To guarantee stable behaviour, the baseline must in general match the smallest negative reward (e.g., $R'(\tau) = R(\tau) - c \ge 0$).
At this point, however, the baseline effectively removes negative trajectories from the objective function -- losing the information contained in these trajectories.
Importantly, the use of importance sampling avoids this issue. To see this, we start by noting that Eq.~\ref{eq:topr_grad_general} is the gradient of
\begin{align}
    J_{\topr}(\pi) &= \sum_{\tau \in T^+}\mu(\tau)\rho\left(\frac{\pi(\tau)}{\mu(\tau)}, a^+, b^+\right) R(\tau) + \sum_{\tau \in T^-} \mu(\tau)\rho\left(\frac{\pi(\tau)}{\mu(\tau)}, a^-, b^-\right) R(\tau),\label{eqn:topr_loss}
\end{align}
where $\rho(\cdot, a, b) : [0, \infty) \to \bR$ is the \emph{taper function}
\begin{equation*}
    \rho(x, a, b) = \begin{cases}
    a \left(1 + \log \frac{x}{a}\right)&\text{if}\ x < a \\
    b\left(1 + \log \frac{x}{b}\right) &\text{if}\ x > b\\
      x &\text{otherwise}.
      \end{cases}
\end{equation*}
The taper function $\rho$ describes the effect of the truncation on the objective optimized by TOPR. It defines a lower bound on the importance ratio, in the sense that for any $a \le b$,
\begin{equation*}
    \rho\left(\frac{\pi(\tau)}{\mu(\tau)}, a, b\right) \le \frac{\pi(\tau)}{\mu(\tau)}\, ;
\end{equation*}
and it is equal to this ratio on the interval $[a, b]$ (Fig.~\ref{fig:taper_function}).
For our canonical choice of $a^+ = b^+ = 1$ a positive reward function $R(\tau) \ge 0)$, this implies that TOPR optimizes a lower bound on the true objective $J(\pi)$:
\begin{align*}
    J(\pi) \geq J_{\topr}(\pi) = \expect_{\tau \sim \mu} \left[ \rho \left ( \frac{\pi(\tau)}{\mu(\tau)}, 1, 1 \right ) R(\tau)\right] \; .
\end{align*}
The bound follows from the analysis of related algorithms by~\citet{deisenroth2013survey,leroux2016efficient,leroux2017tighter} and \citet{gulcehre2023reinforced}.
The following proposition establishes the stable off-policy behaviour of TOPR for a wider range of truncation parameters.
\begin{proposition}\label{prop:bounded_objective}
    For $a^- = 0$, Equation~\ref{eqn:topr_loss} is bounded above: there exists a $B$ such that
    \begin{equation*}
        \sup\nolimits_\pi J_{\topr}(\pi) \le B.
    \end{equation*}
    Furthermore, for any $a^- > 0$, $J_{\topr}(\pi)$ is unbounded above unless $R(\tau) \ge 0$ for all $\tau$.
\end{proposition}

For positive examples, the taper function (with $a^+ = 1$) maintains a substantial gradient even when $\pi(\tau)$ is small. This is because
 the weight used for the gradient of the surrogate objective, $\mu(\tau)$, is independent of the current policy $\pi$. This make it possible for the model to recover from a low probability $\pi(\tau)$ assigned to a good trajectory, avoiding a traditional failure of REINFORCE. Empirically, \citet{leroux2017tighter} observed lower variance and more efficient learning when optimizing this log-ratio lower bound. With negative rewards, however replacing the importance ratio with the log-ratio leads to the surrogate objective being an \emph{upper bound} on $J(\pi)$~\citep{leroux2016efficient}, which is a different way of expressing the conclusion of Proposition~\ref{prop:bounded_objective}.

\subsection{The importance of dataset composition}\label{sec:dataset_composition}
In addition to the choice of the loss, the composition of the training set is critical to the performance of the trained model. In the context of training language models to perform chain-of-thought reasoning, for example, dataset curation methods such as STaR, ReST, and ReST-EM differ mainly on which data they include.

As we will see, the relative importance of positive and negative examples in the dataset is equally critical to good performance. Interestingly enough, the baseline parameter can also be interpreted as modulating this relative importance.

Let us again assume a binary reward function $R_0(\tau) \in \{-1,1\}$ and a baseline $c \in [-1, 1]$. Let $p = |T^+|/(|T^+| + |T^-|)$ be the proportion of positive examples in the dataset. Substituting $R(\tau) = R_0(\tau) - c$ into Eq.~\ref{eqn:topr_loss}, we obtain
\begin{align*}
J_{\topr}(\pi) &= \sum_{\tau \in T^+}\mu(\tau)\rho\left(\frac{\pi(\tau)}{\mu(\tau)}, a^+, b^+\right) (1-c) + \sum_{\tau \in T^-} \mu(\tau)\rho\left(\frac{\pi(\tau)}{\mu(\tau)}, a^-, b^-\right) (-1-c)\\
&= (1-c)\sum_{\tau \in T^+}\mu(\tau)\rho\left(\frac{\pi(\tau)}{\mu(\tau)}, a^+, b^+\right) - (1+c)\sum_{\tau \in T^-} \mu(\tau)\rho\left(\frac{\pi(\tau)}{\mu(\tau)}, a^-, b^-\right) \; .
\end{align*}
With this transformation, the contribution of each positive example to the objective is weighted by $1-c$. With some algebra, we find that the \emph{effective} proportion of positive examples changes from $p$ to
\begin{equation}\label{eqn:effective_proportion}
    \tilde p = \frac{p(1-c)}{p(1-c) + (1-p)(1+c)} = \frac{p(1-c)}{1+(1-2p)c} .
\end{equation}
With a fixed dataset, we can thus vary the relative importance of positive and negative examples ($\tilde p$ and $1- \tilde p$) by modifying the baseline according to Equation~\ref{eqn:effective_proportion}.
This generalizes the result from the previous section that discarding negative examples is equivalent to using a baseline of $-1$.

Furthermore, the choice of the baseline $c$ can be viewed as adding a softer version of the $\kl(\mu \cdbar \pi)$ term to the TOPR objective, again encouraging $\pi$ to stay close to $\mu$ when $c < 0$ (see Appendix~\ref{sec:appendix_baselines}). As a negative baseline $c$ also increases the effective proportion of positive examples in the dataset, we see that a larger proportion of positive examples will decrease the degree of off-policyness that is achievable without resampling the training set. Adding negative examples can thus be seen as a way to further improve the policy.
We shall show in Section~\ref{sec:results} how carefully choosing the effective proportion of positive examples, either through dataset composition or a baseline, can lead to a boost in accuracy.

\section{Results}
\label{sec:results}
We study the effectiveness of TOPR at training language models to perform one of two tasks: chain-of-thought (CoT) reasoning and verifying such a reasoning. For the most part we focus on the single-iteration, fully offline regime aiming to characterize the relative stability and effectiveness of TOPR for training language models compared to prior alternatives. Our results are naturally complementary to the full gamut of methods that improve language models iteratively.

\subsection{Models and Datasets}
\label{sec:appendix:datasets}
We focus on mathematical reasoning datasets that require step-by-step solutions and are widely used to evaluate the reasoning capabilities of LLMs. As our core model, we use the Llama 3 family of instruction-tuned language models \citep{dubey2024llama}, using the 8B model unless otherwise specified.

\paragraph{GSM8K \citep{gsm8k}}
The GSM8K dataset is composed of short grade-school math problems, requiring basic arithmetic or elementary algebra to solve. It contains 1,319 problems for testing and 7,473 for training. Verifying the correctness of model responses is straightforward, as the final answer is typically an integer. When the string is not present, we consider the answer as missing.
For each training question we generate $n=16$ candidate solutions using chain-of-thought (CoT) prompting, using the 8-shot prompt from \citet{wang2022self}.
We parse our model's answer by looking for the magic string ``The answer is'', matching this few-shot prompt. 

\paragraph{MATH \citep{MATH}}
The MATH dataset contains problems from high school math competitions, covering a wide range of topics such as algebra, geometry, and probability, and is generally harder than GSM8K. We use the split from the original work, which includes 7,500 training problems. For computational reasons, we report performance on the smaller MATH-500 test set \citep{lightman2023let}.
Each problem includes a step-by-step solution, ending in a final answer marked by \verb+\boxed{}+
in the solution (e.g., ``\emph{..so the smallest possible value of $c$ is $\boxed{\pi}$''}). 
This marking allows for verification of the correctness of model-generated responses by comparing the final answer to the ground truth. We use the script provided by \citet{deepseek-math} for this purpose.

For each training question we generate $n=32$ candidate solutions using chain-of-thought (CoT) prompting, using the 4-shot prompt from \citet{lewkowycz2022solving}.

\subsection{Experimental setup}

Our training infrastructure is based on HuggingFace's transformers library \citep{wolf20transformers}. We use data parallelism on a single H100 node with a per-GPU batch size of 1, a constant learning rate of 5e-7 chosen from a small parameter sweep, the Adafactor optimizer \citep{shazeer18adafactor} to minimize memory usage, and neither weight decay nor KL regularization.\footnote{In addition to allowing us to focus on the relative stability of different learning rules, the removal of the KL term decreases the memory and computational burden of training the language model.}
We divide the loss by the sequence length~\citep{grinsztajn2024averaging}, which in reinforcement learning terms can be thought of as implementing hyperbolic discounting. We use HuggingFace's default gradient clipping parameter of 1.0. The reward is +1 for a correct answer and -1 for an incorrect answer, and no baseline is used. Candidate generations are produced using vLLM \citep{kwon2023efficient} with temperature $T=1$, $\textrm{top}_p=1$, $\textrm{top}_k=500$, and a maximum of 512 tokens.

An iteration of training consists of generating candidates using a model (usually the base model), labelling those candidates with their associated reward, and performing a single epoch over this generated dataset. The reference $\mu$ corresponds to the model predictions at the beginning of the iteration.
Unless otherwise noted, all reported scores and accuracies are with respect to test sets, measured by evaluating the model at the end of the iteration. No early stopping is performed.
We use the bootstrap technique \citep{efron1994introduction} to provide confidence intervals: we generate 64 solutions for each question for GSM8K and 16 for MATH due to the cost of evaluating. For each question, we select K answers at random without replacement, then compute the average maj@K or pass@K performance across the dataset. We repeat this process 100 times and estimated the empirical variance $\hat{V}$ across the 100 trials.
We compute the standard error as $\sqrt{\frac{V}{\frac{64}{K} - 1}}=\sqrt{\frac{KV}{64-K}}$; depicted confidence intervals correspond to two standard errors.

\subsection{Fine-tuning chain-of-thought reasoning}

\begin{figure}
    \centering
    \includegraphics[width=.495\linewidth]{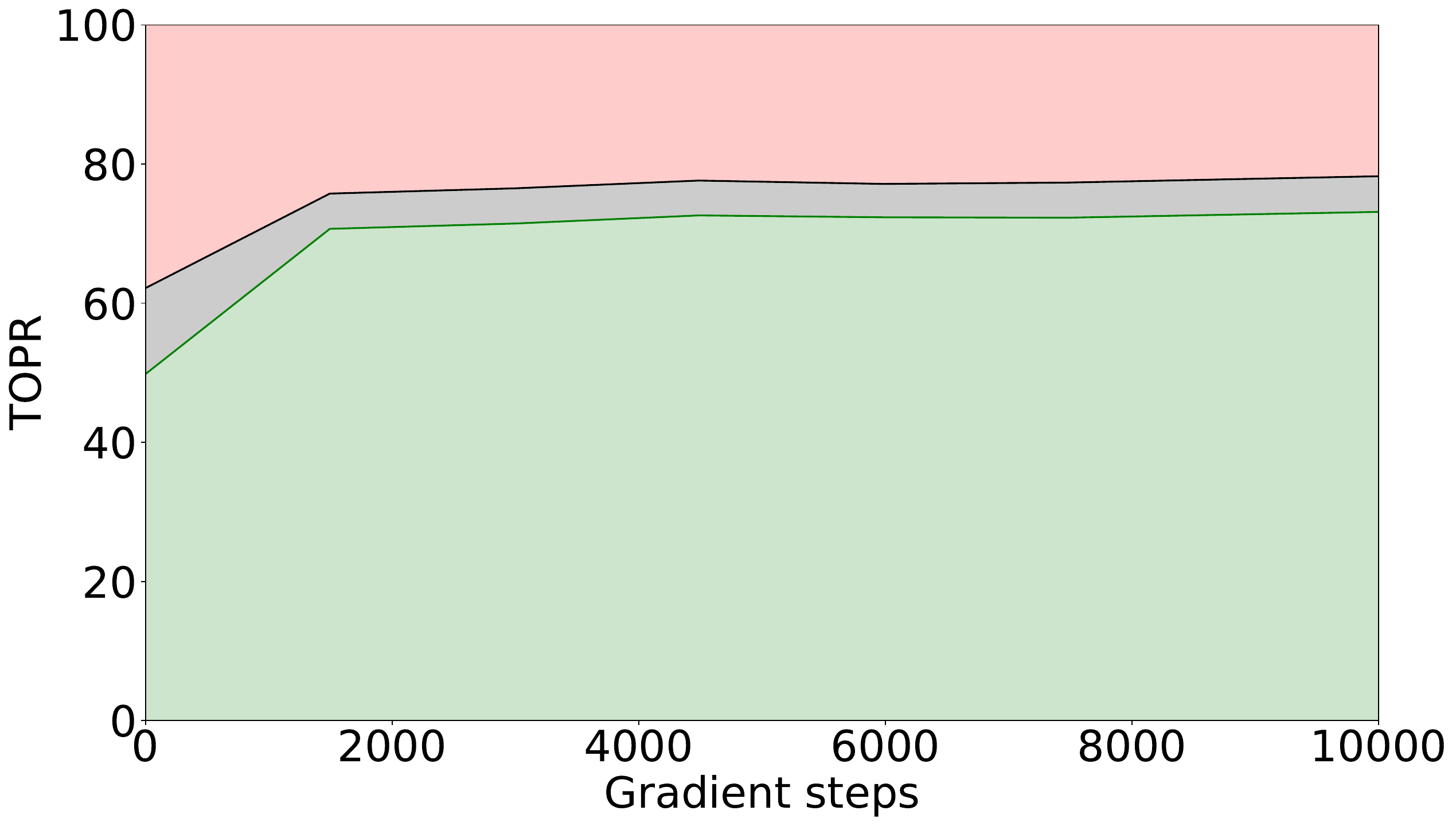}
    \includegraphics[width=.495\linewidth]{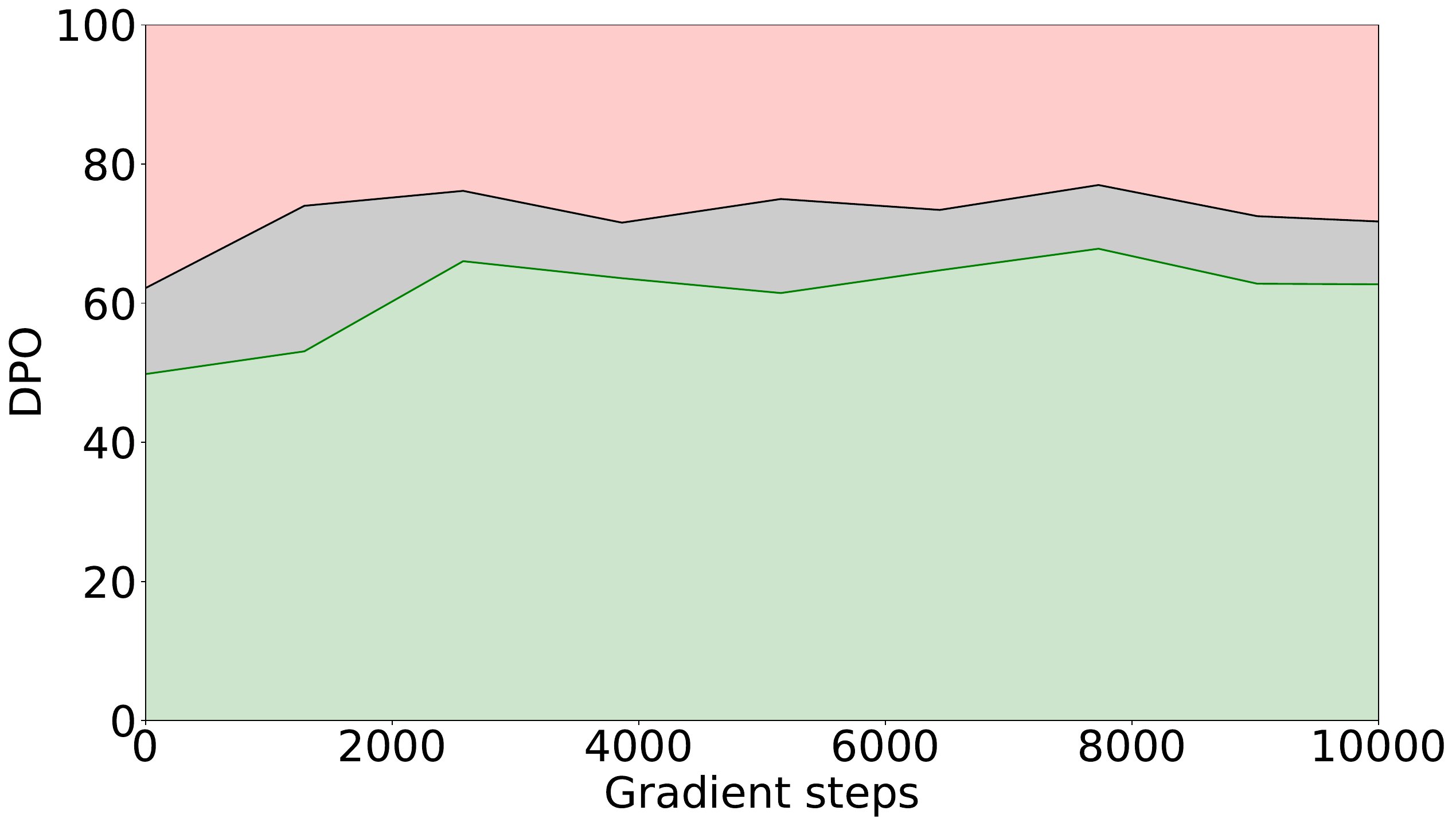}
    \includegraphics[width=.495\linewidth]{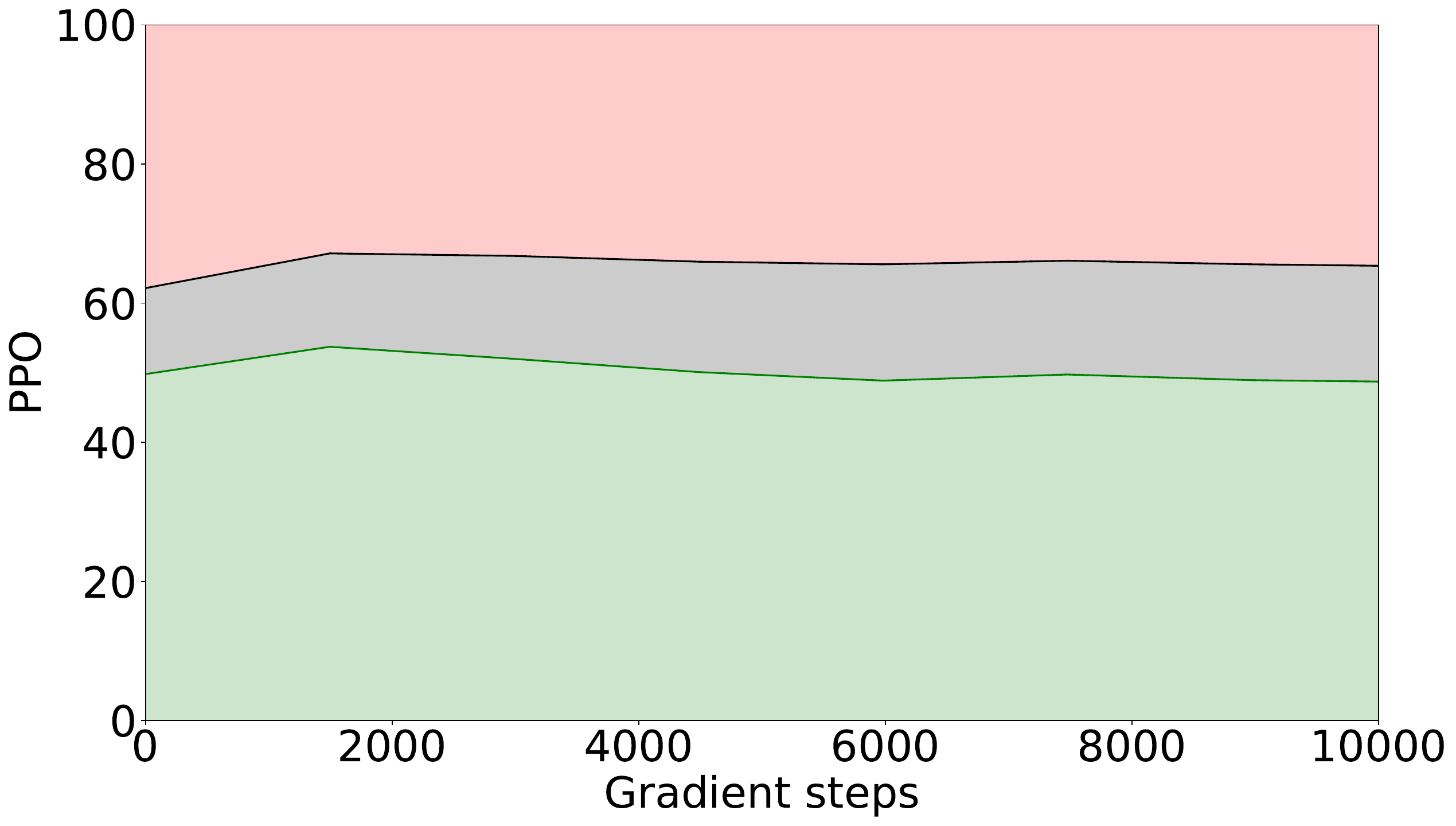}
    \includegraphics[width=.495\linewidth]{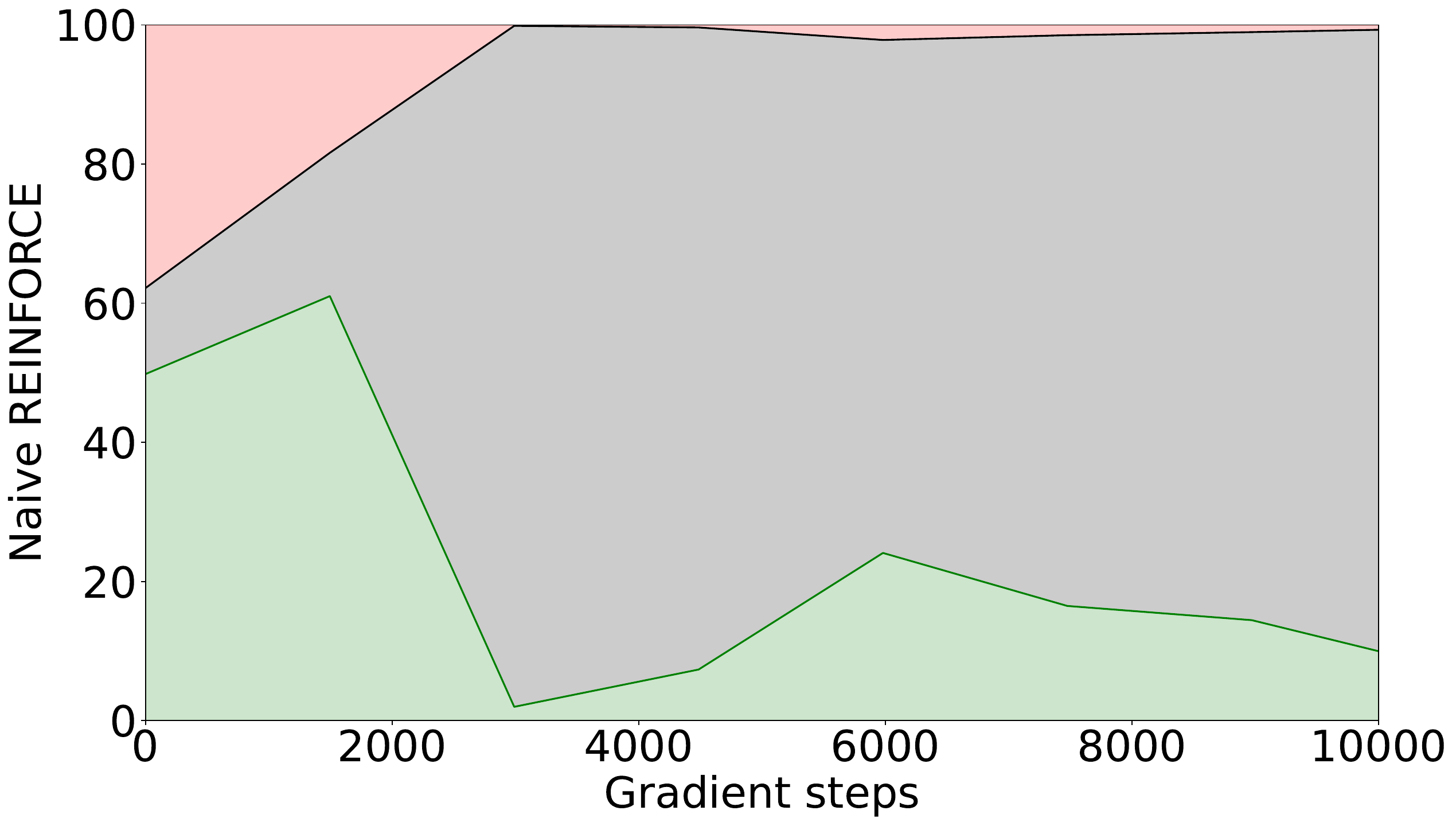}
    \caption{\textbf{Proportion of correct (green), incorrect (gray), and invalid (red) generated solutions on the GSM8K test set.} Of the four, TOPR is the only one to significantly reduce invalid generations.
    \label{fig:gsm8k_G8b_with_bad_answers}}
\end{figure}

Our first set of experiments aims to answer the question: \emph{Is a more careful handling of importance ratios and rewards beneficial in off-policy policy optimization?} We begin with a comparison of PPO, DPO, naive REINFORCE, and TOPR.
For DPO, pairs of candidate solutions are formed from these so as to obtain up to $16$ contrastive pairs.
For PPO, we use $\epsilon = 0.2$.

Fig.~\ref{fig:gsm8k_G8b_SC_DPO+PPO+TOPR+Naive} experimentally demonstrates the limitations of existing methods. Naive REINFORCE performs well at first but, in the absence of a KL term, collapses as $\pi$ moves away from $\mu$. As expected, PPO makes little progress on the objective as most data points quickly fall outside of its $[1-\epsilon, 1+\epsilon]$ range. DPO performs well off-policy, confirming the observations of~\citet{noukhovitch2024asynchronous}, especially when measured in terms of pass@1 accuracy.

By contrast, TOPR rapidly improves on the base model, eventually reaching a respectable level of accuracy.

\paragraph{TOPR minimizes reasoning failures.} 
To understand the reasons behind TOPR's success, we measured the proportion of generated solutions that were correct, incorrect, or invalid (the string ``The answer is'' is not present) during the course of training.
Fig.~\ref{fig:gsm8k_G8b_with_bad_answers} gives strong evidence as to the root cause of REINFORCE's poor performance, whose generations are overwhelmingly degenerate by the end of training. By contrast, TOPR proves effective at teaching the model to avoid incorrect formatting -- yielding the desirable property that one can solely rely on RL for solution generation, rather than using additional tools to correctly format them.

\begin{figure}
    \centering
    \hspace{1.3em}
    \includegraphics[width=.475\linewidth]{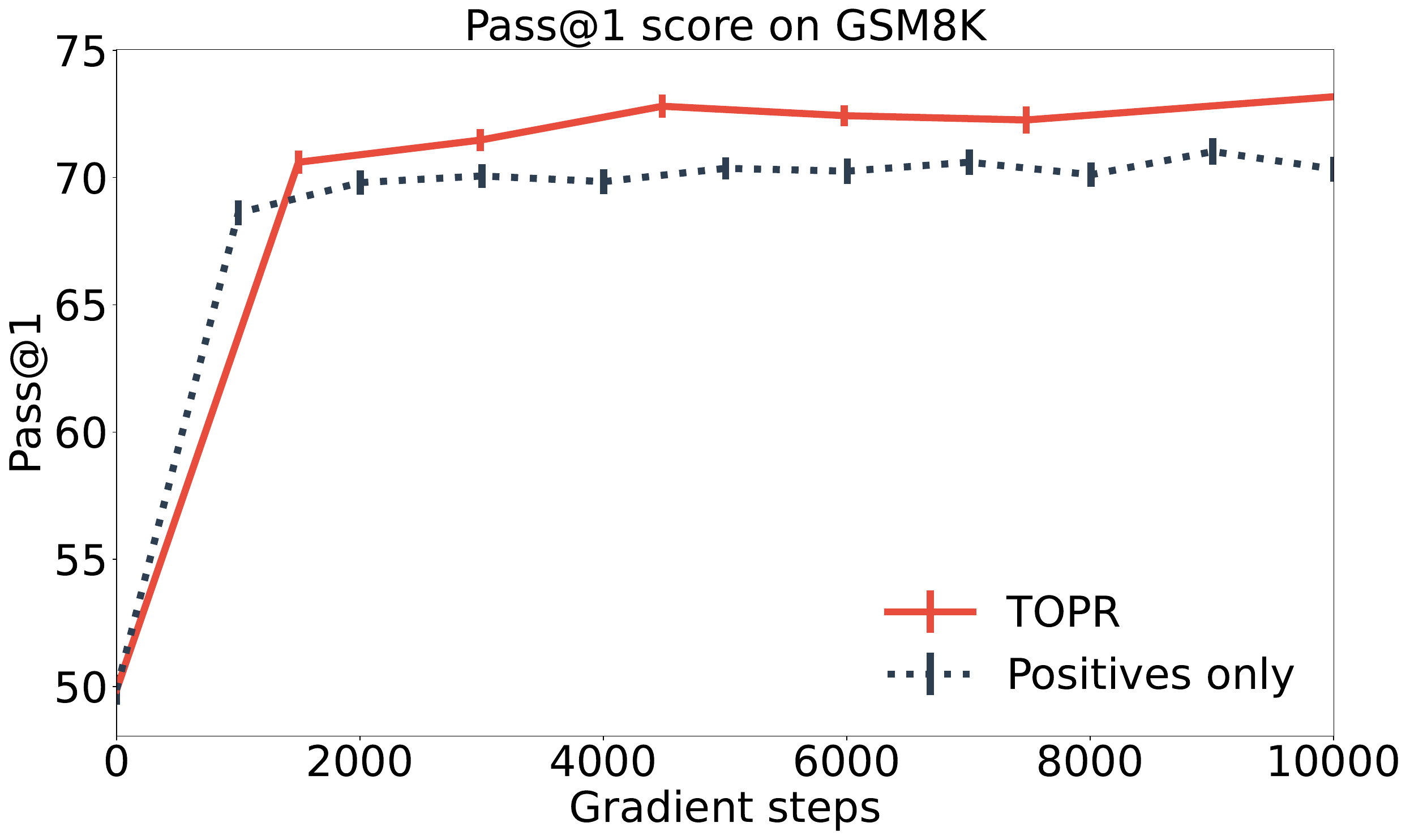}
    \hspace{0.2em}
    \includegraphics[width=.475\linewidth]{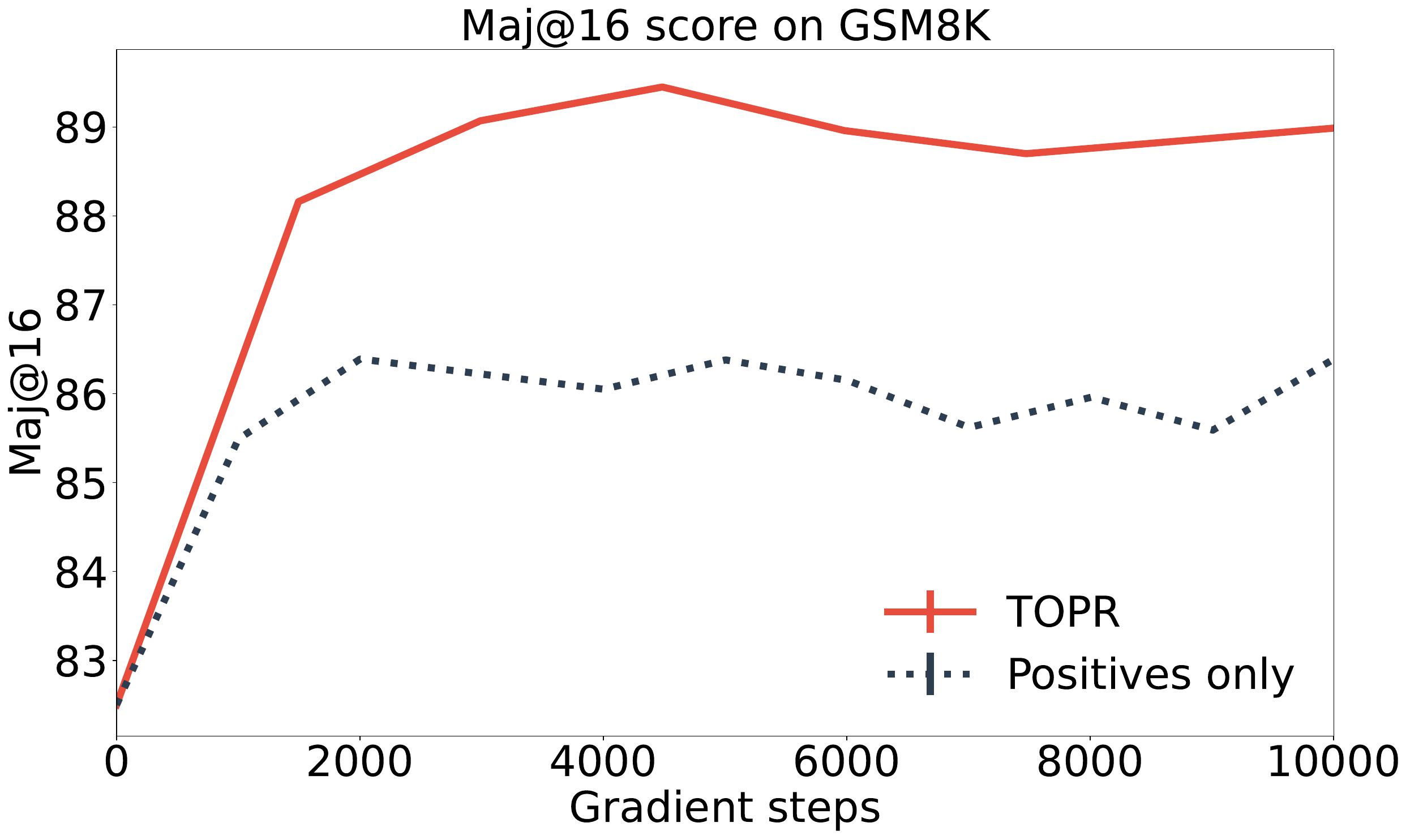}
    \vspace{1em}
    \includegraphics[width=.98\linewidth]{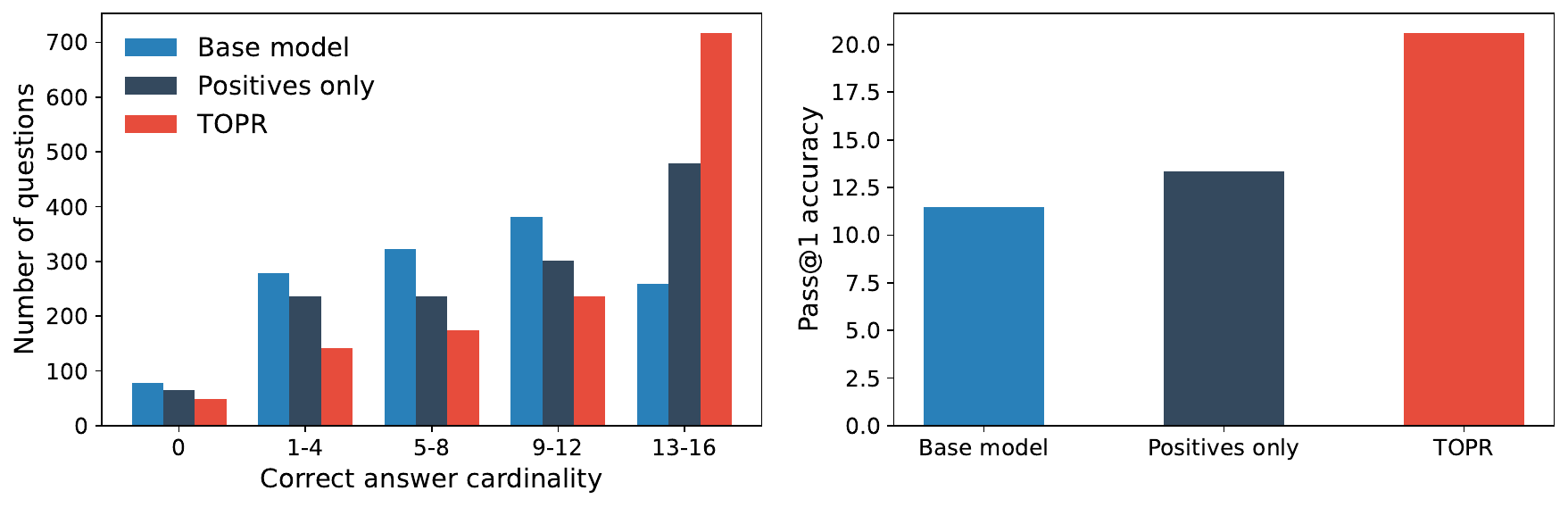}
    \caption{\textbf{Top} Test set accuracy on GSM8K across training when using all examples (TOPR) or positive examples alone. Not shown here, TOPR also yields higher inference efficiency at test time: greater maj@n performance for all n. \textbf{Bottom left:} Distribution of GSM8K problem questions as a function of the number of correct generations. TOPR more effectively reduces the number of questions with none (0) to few (1--4) correct generations. \textbf{Bottom right:} Pass@1 accuracy on MATH.
    \label{fig:gsm8k_G8b_SC_TOPR+Positives}}
\end{figure}

\paragraph{Using negative examples improve performance.} To understand the impact of negative examples on training, we next formed a ``positives only'' dataset by removing all negative examples from our base dataset. This procedure mimics some of the design choices of recent work such as STaR that apply SFT as the inner loop of an RL-like procedure.
Using this dataset results in stable learning but substantially lower performance than that of TOPR (Fig.~\ref{fig:gsm8k_G8b_SC_TOPR+Positives}, top).

This translates into greater self-consistency~\citep{wang2022self} efficiency: more solutions must be generated at test time to reach the same level of performance (top right). Breaking down the test results as a function of the number of rationales that conclude in the correct answer (``Correct answer cardinality'', bottom left),
we find that TOPR's performance gains from using negative examples can be attributed to reducing the number of questions for which no or few solutions are found, guaranteeing a strong majority for self-consistency. We find similar results on MATH, where using TOPR enables us to almost double the pass@1 accuracy compared to the base model (bottom right). Beyond these results, it is also worthwhile noting that TOPR is more \emph{training inference-efficient}: indeed, because all vLLM generations are used to improve the model, training data is effectively generated as a faster rate.

\paragraph{Striking the right balance of positive and negative examples.}
We now refine the previous analysis by varying the effective proportion of positive examples in the dataset. We do this by selecting two datasets containing 50,000 examples: one with 10\% of positive examples (labelled ``10p'') and one with 50\% of positive examples (``50p''). We then vary the baseline for each model to reach an effective proportion of positive examples from 1\% all the way to 100\% (achieved by setting the baseline to $c=-1$).

\begin{figure}
    \centering
    \includegraphics[width=0.495\linewidth]{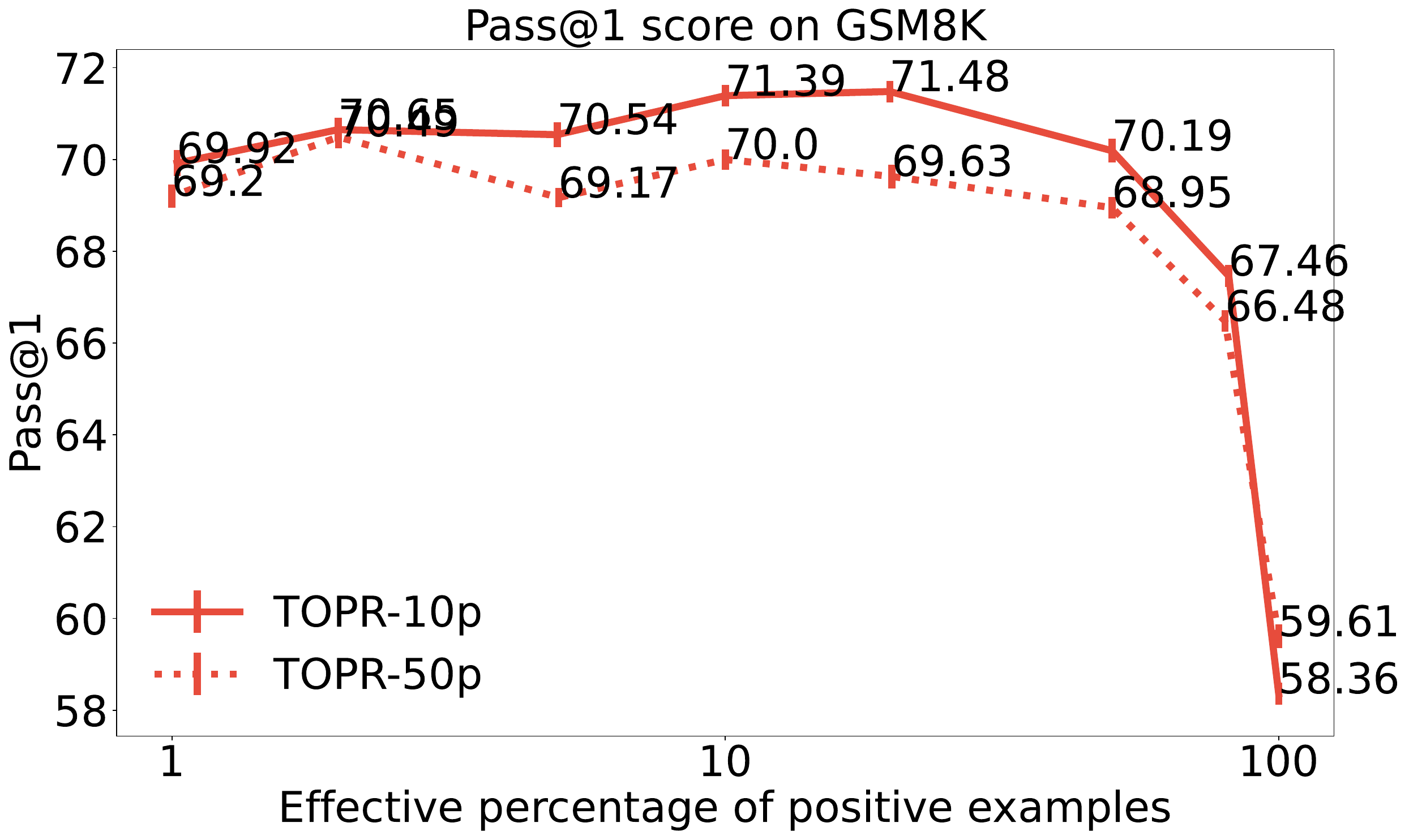}
    \includegraphics[width=0.495\linewidth]{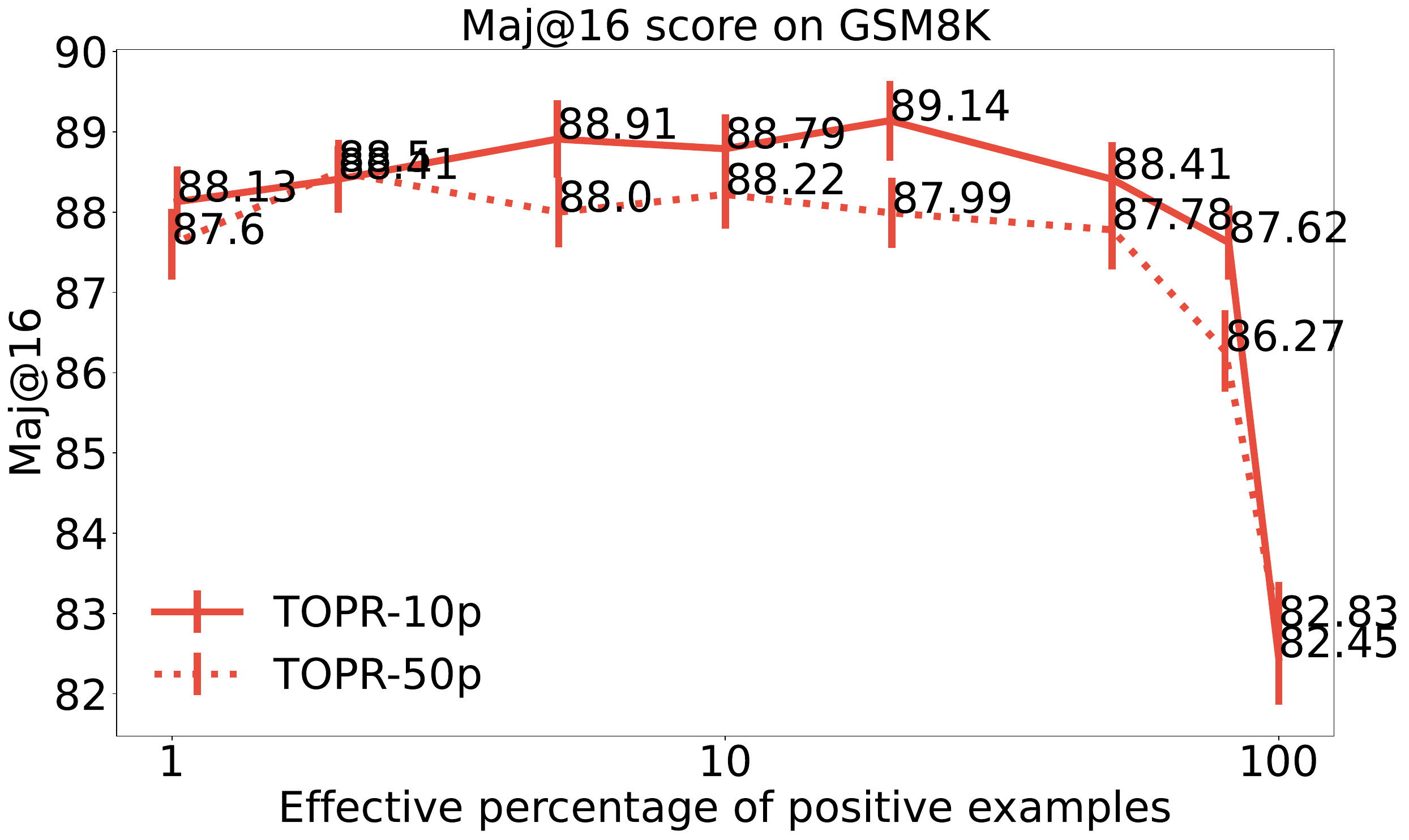}
    \caption{\textbf{Test set accuracy on the GSM8K dataset when the training set contains either $p=10\%$ (solid line) or $p=50\%$ (dotted line) of positive examples as the baseline $c$ is varied.} The x-axis, on a log scale, represents the effective proportion of positives $\tilde p = \frac{p(1-c)}{1+c(1-2p)}$.}
    \label{fig:TOPR_baseline_proportion}
\end{figure}

\begin{figure}
    \centering
    \includegraphics[width=0.495\linewidth]{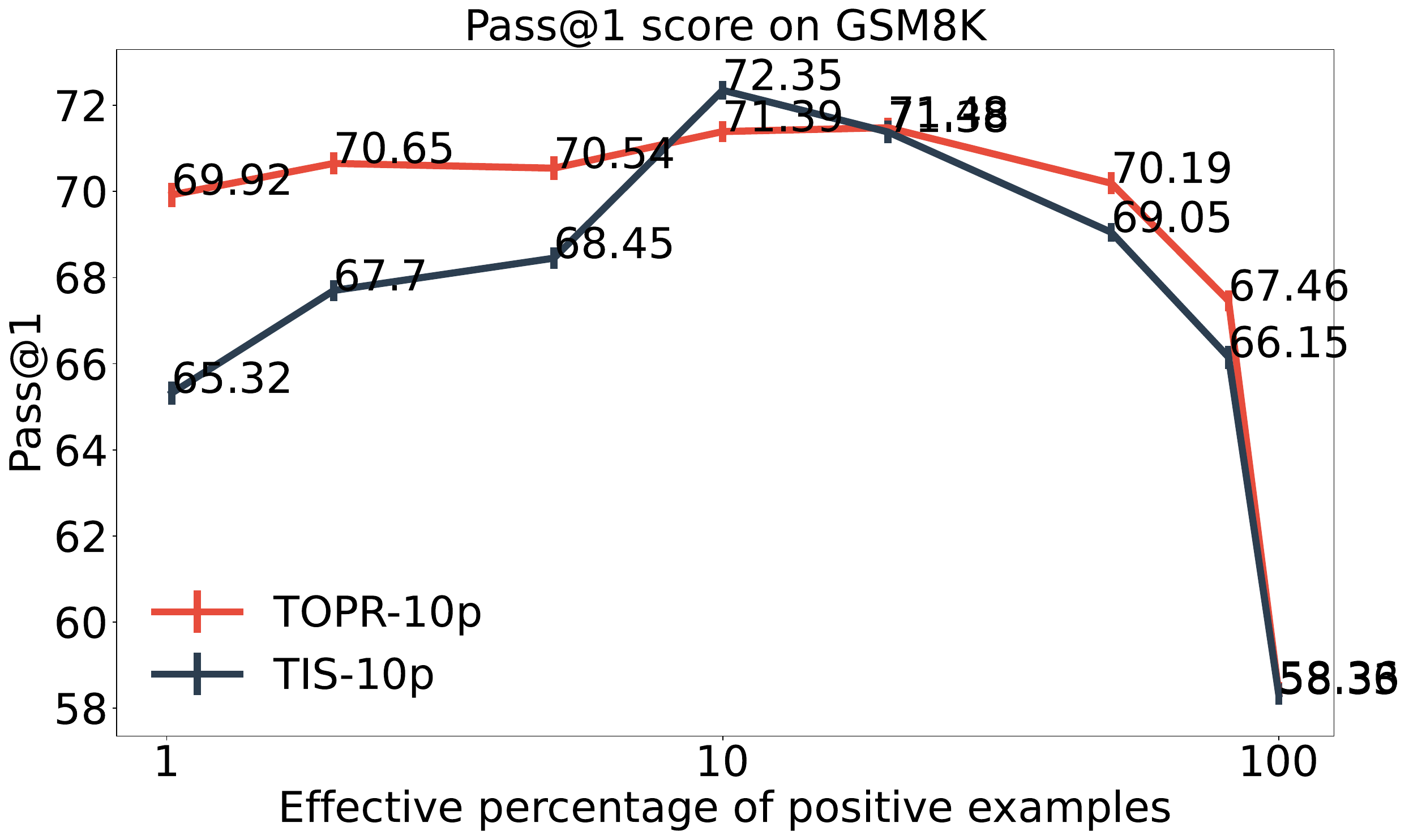}
    \includegraphics[width=0.495\linewidth]{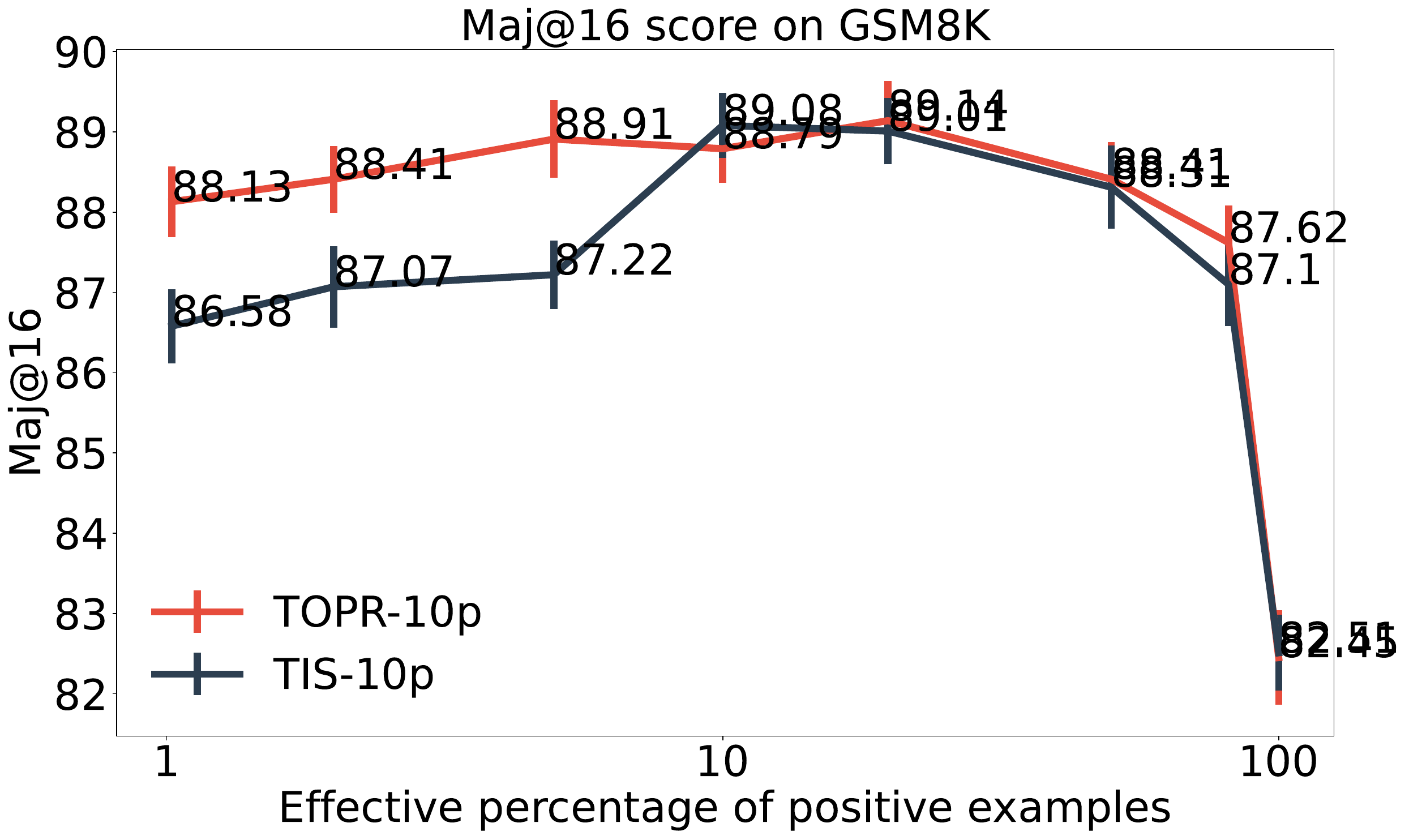}
    \includegraphics[width=0.495\linewidth]{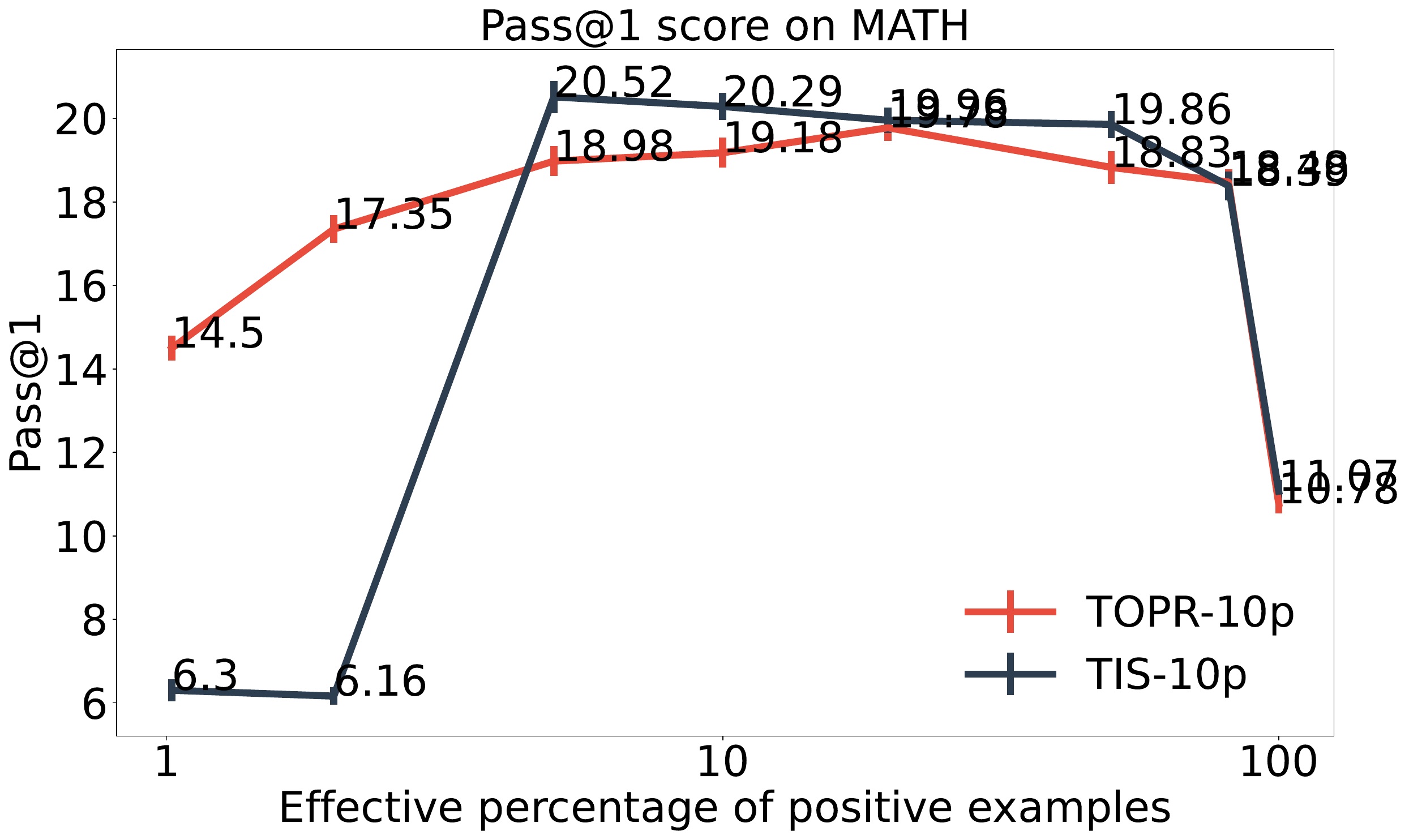}
    \includegraphics[width=0.495\linewidth]{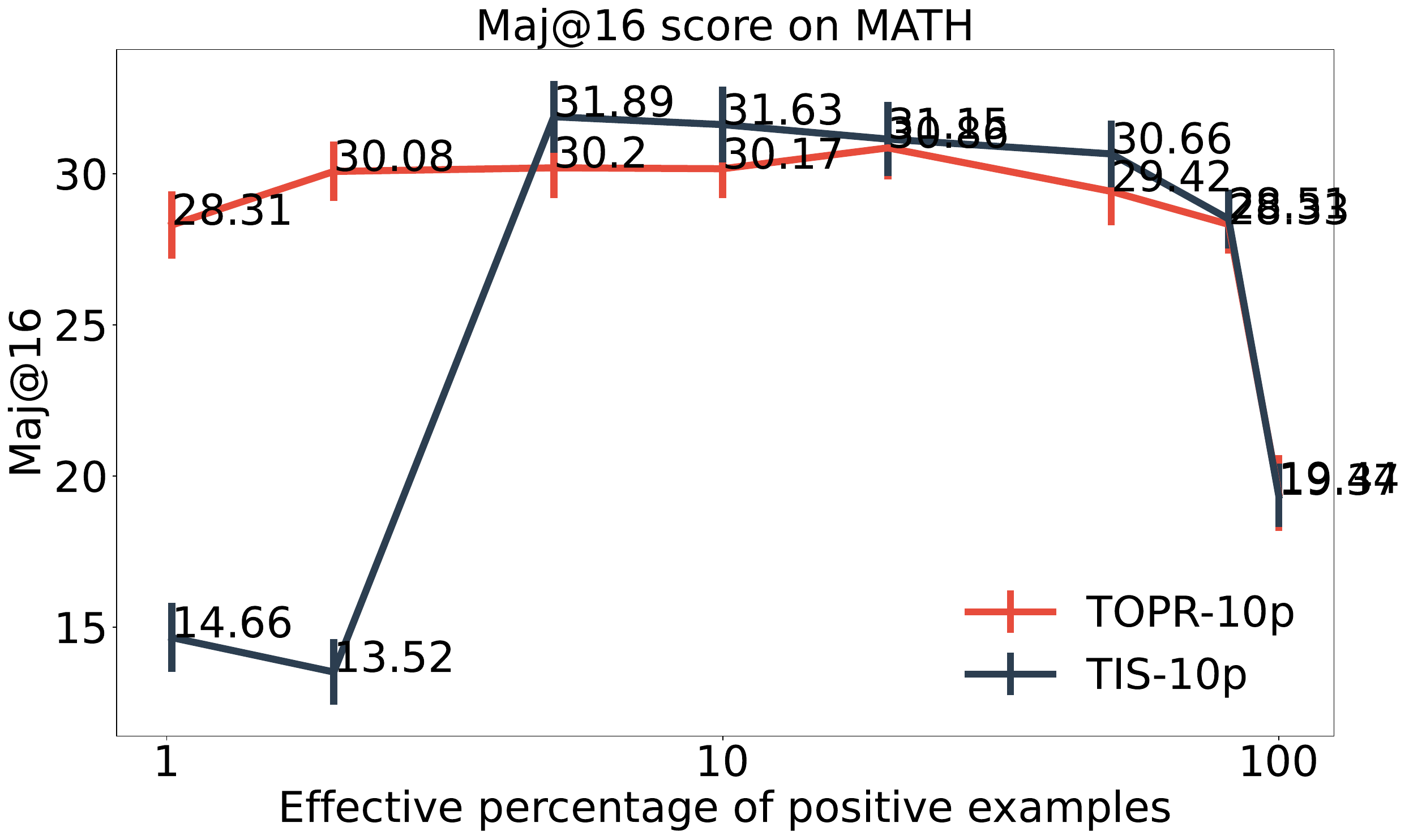}\caption{\textbf{Test set accuracy on GSM8K (top) and MATH (bottom) when the training set contains $p=10\%$ of positive examples, using either truncated importance sampling (TIS) or TOPR.} The x-axis, on a log scale, represents the effective proportion of positives $\tilde p = \frac{p(1-c)}{1+c(1-2p)}$.}
    \label{fig:TOPR+TIS_baseline_proportion}
\end{figure}

Figure~\ref{fig:TOPR_baseline_proportion} shows that TOPR's performance is maximal around 10-20\% of effective positive examples, regardless of the \emph{actual} proportion of positive samples in the training - 10\% for the solid curve, 50\% for the dashed curve. The performance drops if the proportion is either too small or too large, and markedly decreases as the proportion of positive examples goes above 50\%. We attempt to explain this fact from an optimization perspective in Appendix~\ref{sec:appendix_negatives}, from a regularization perspective in Appendix~\ref{sec:appendix_baselines}, and from a generalization perspective in Appendix~\ref{sec:appendix_rl_supervised}.
We observe a strong correlation between the 10\% and 50\% curves, showing that the effective proportion is a more critical factor than the actual proportion of positive samples. Further, Fig.~\ref{fig:TOPR+TIS_baseline_proportion} shows the optimal effective proportion to be around 10-20\% for both GSM8K and MATH. We posit that a good baseline is one that achieves such a proportion.

Our result also gives further evidence that the optimal baseline is not always the expected return in practical settings, contrary to common belief.\footnote{\citet{peters08reinforcement} and \citet{deisenroth2013survey} remark on similar findings in the on-policy setting; see also the empirical study by \citet{chung2021beyond}.}
Finally, we see that the model reaches slightly, but consistently higher performance when the percentage of positives is obtained through sampling rather than by setting a baseline. We will demonstrate just below how to leverage this characteristic to improve training performance.

\paragraph{Acceleration improves robustness to dataset composition.} Given the important performance gains from incorporating negative examples to training and the relevance of dataset composition (Section~\ref{sec:dataset_composition}), it is natural to ask whether TOPR's positive-example acceleration ($a^+ = 1$) helps when positive examples are outnumbered in the dataset, for example because the problem at hand is very hard. Figure~\ref{fig:TOPR+TIS_baseline_proportion} shows the test performance on GSM8K using either TOPR or TIS when the effective proportion of positive examples varies. When that proportion is low, the model tends to lower the probability of most trajectories in its training set. This leads to the probability of positive trajectories being lowered as well. Thanks to its acceleration ($a^+=1$), TOPR recovers from these cases while TIS cannot. When the effective proportion of positive examples is high, there is virtually no difference between TOPR and TIS. Interestingly, we see that TIS reaches a slightly higher maximum pass@1 accuracy (chosen over all experiments) compared to TOPR. This suggests that TIS may trade robustness for peak performance.

\begin{figure}
    \centering
    \centering
    \begin{minipage}[c]{0.6\textwidth}
        \includegraphics[width=\textwidth]{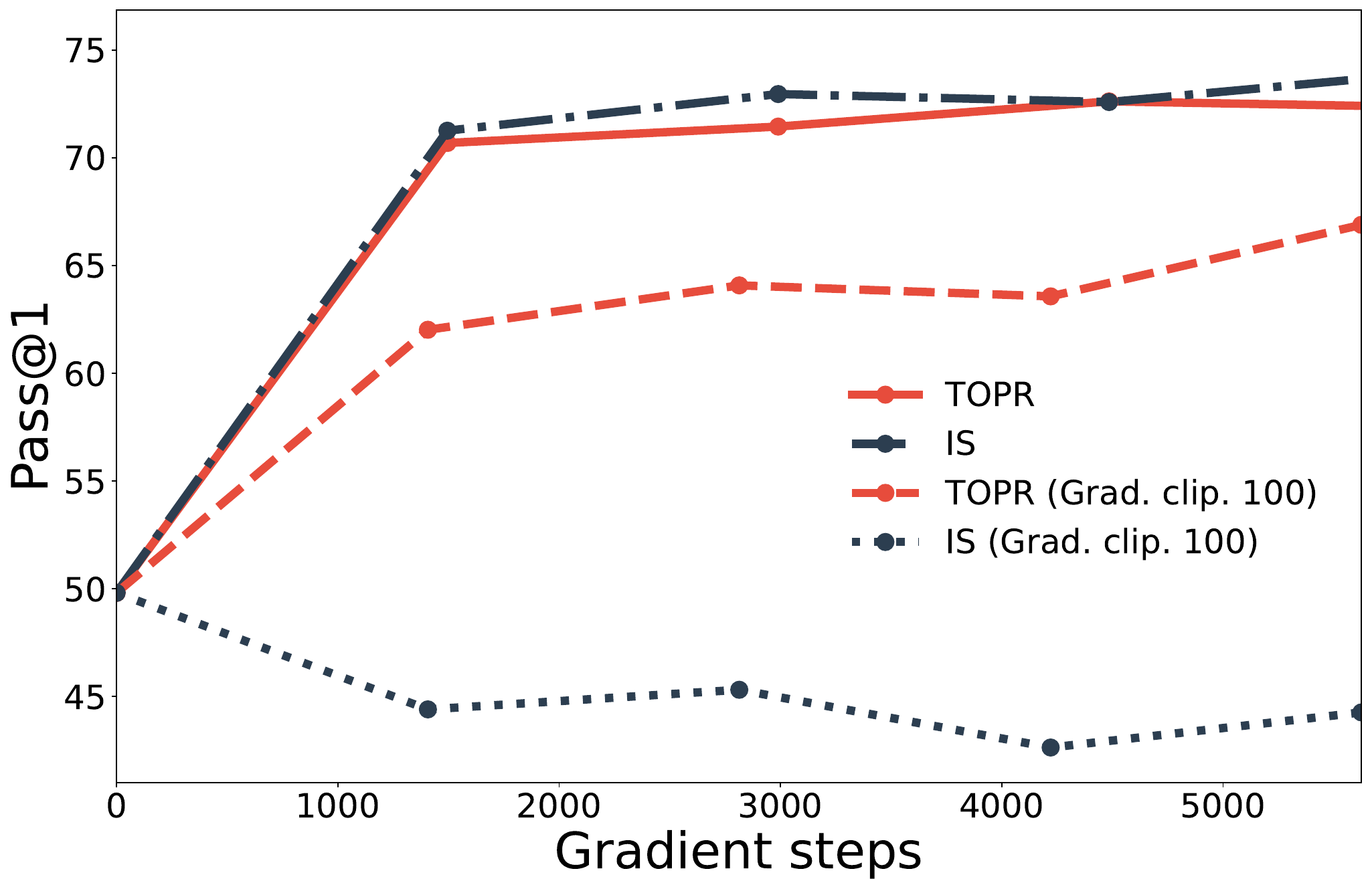}
    \end{minipage}
    \hfill
    \begin{minipage}[c]{0.35\textwidth}
        \caption{\textbf{Performance of TOPR and standard importance sampling (IS)} in our default experimental setting and with a higher gradient clipping parameter (100.0). TOPR shows greater robustness to the gradient clipping parameter.}
    \label{fig:TOPR+TIS_gradient_clipping}
    \end{minipage}
\end{figure}

\paragraph{Ratio truncation improves stability.} A natural question is whether the truncation of importance ratios as done by TOPR is necessary, or if standard importance sampling alone suffices, as suggested by \citet{ahmadian2024back}. To study this question, we trained from the base model as before but using standard importance sampling ($a^+=a^-=0$, $b^+=b^-=+\infty$). 
Fig.~\ref{fig:TOPR+TIS_gradient_clipping} shows that this results in an algorithm that is as performant as TOPR. When we inspected its in-training behaviour, however, we found that the average gradient norm became increasingly larger as training became more and more off-policy. The impact of this norm blow-up is mitigated by the use of the default gradient clipping parameter (1.0), as well as the relatively low number of negative examples in the training dataset ($\sim$33\%). To further demonstrate the stabilizing effect of ratio truncation in TOPR, we used the same two algorithms but now with a negatively-skewed dataset (60\% negatives) and the gradient clipping parameter set to 100.0.  The ``Grad. Clip. 100'' curves depict these results. TOPR is affected by these changes but still improves on the base model. Standard importance sampling, on the other hand, harms the model's performance -- producing \textbf{31\%} of bad reasonings by the end of training against \textbf{12\%} for the base model.

\begin{figure}
    \centering
    \includegraphics[width=.495\linewidth]{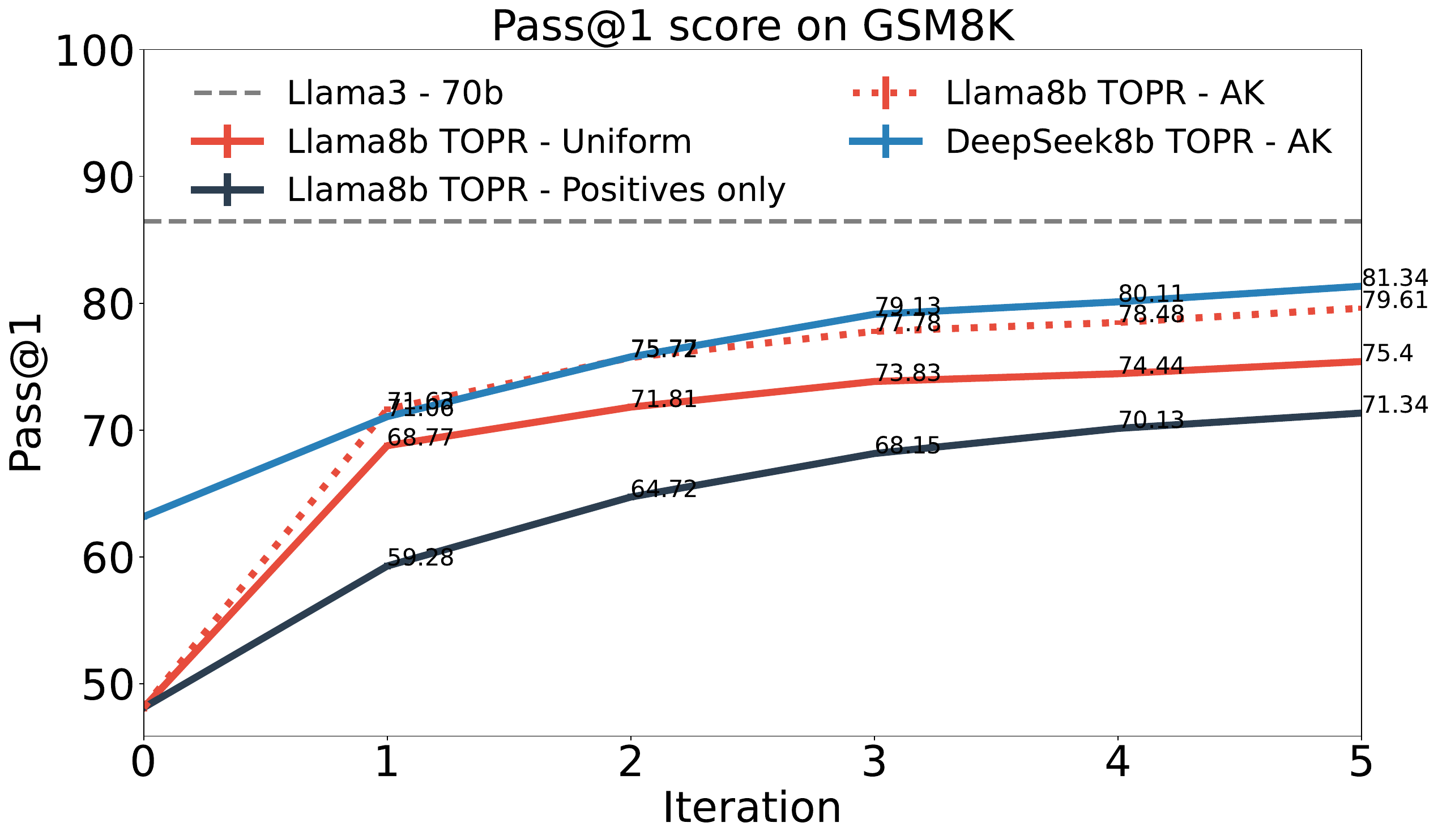}
    \includegraphics[width=.495\linewidth]{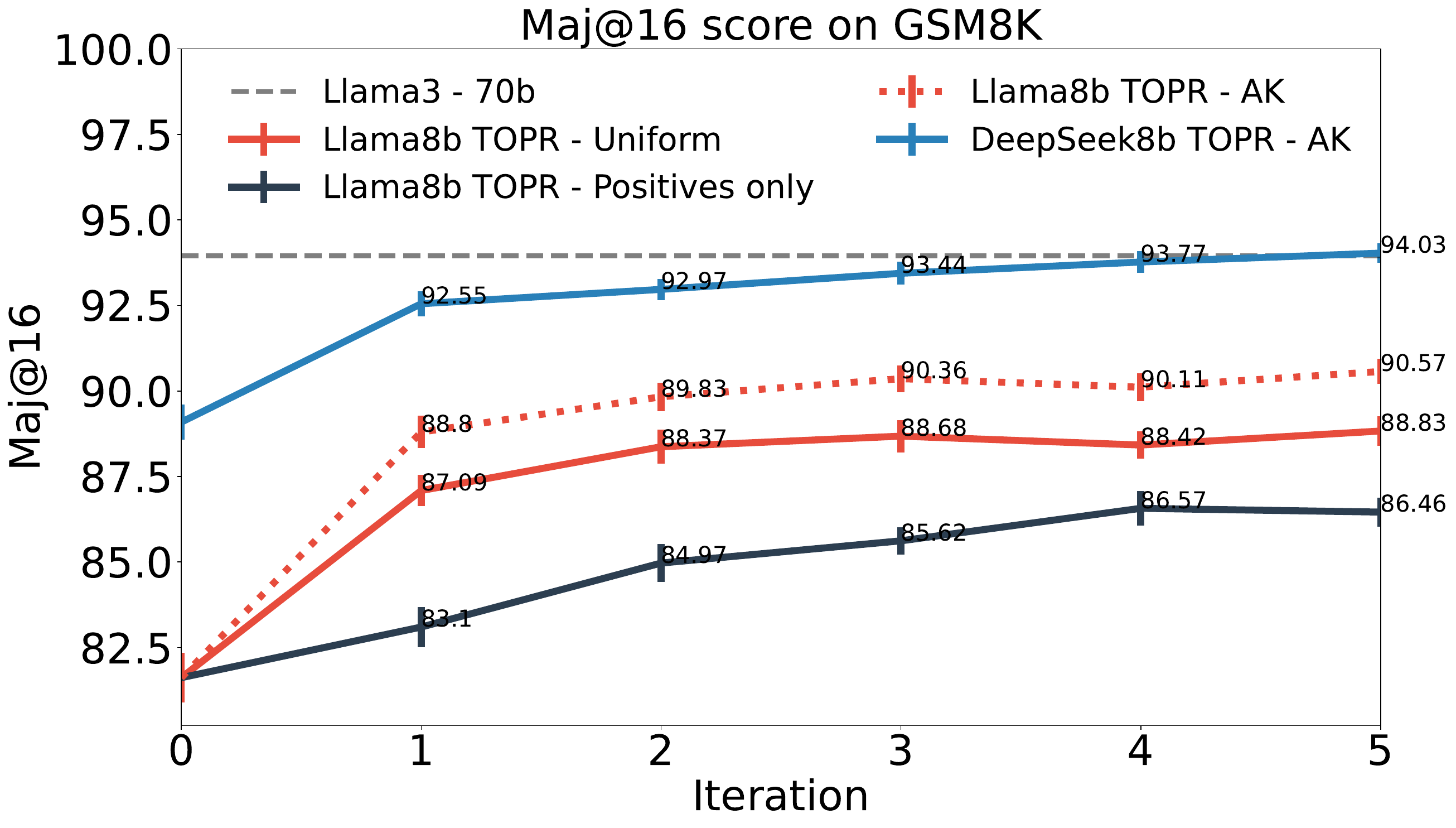} 
    \includegraphics[width=.495\linewidth]{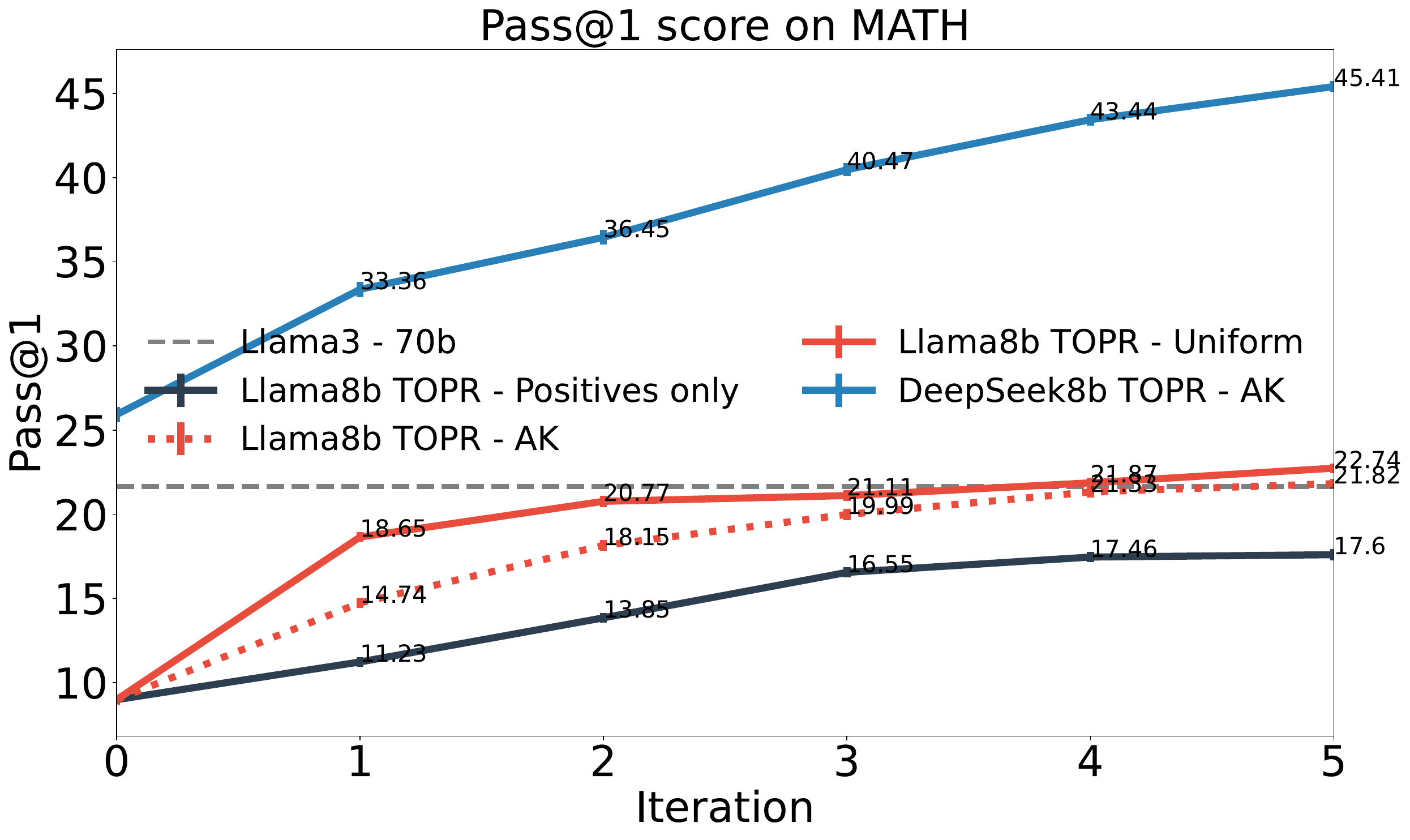}
    \includegraphics[width=.495\linewidth]{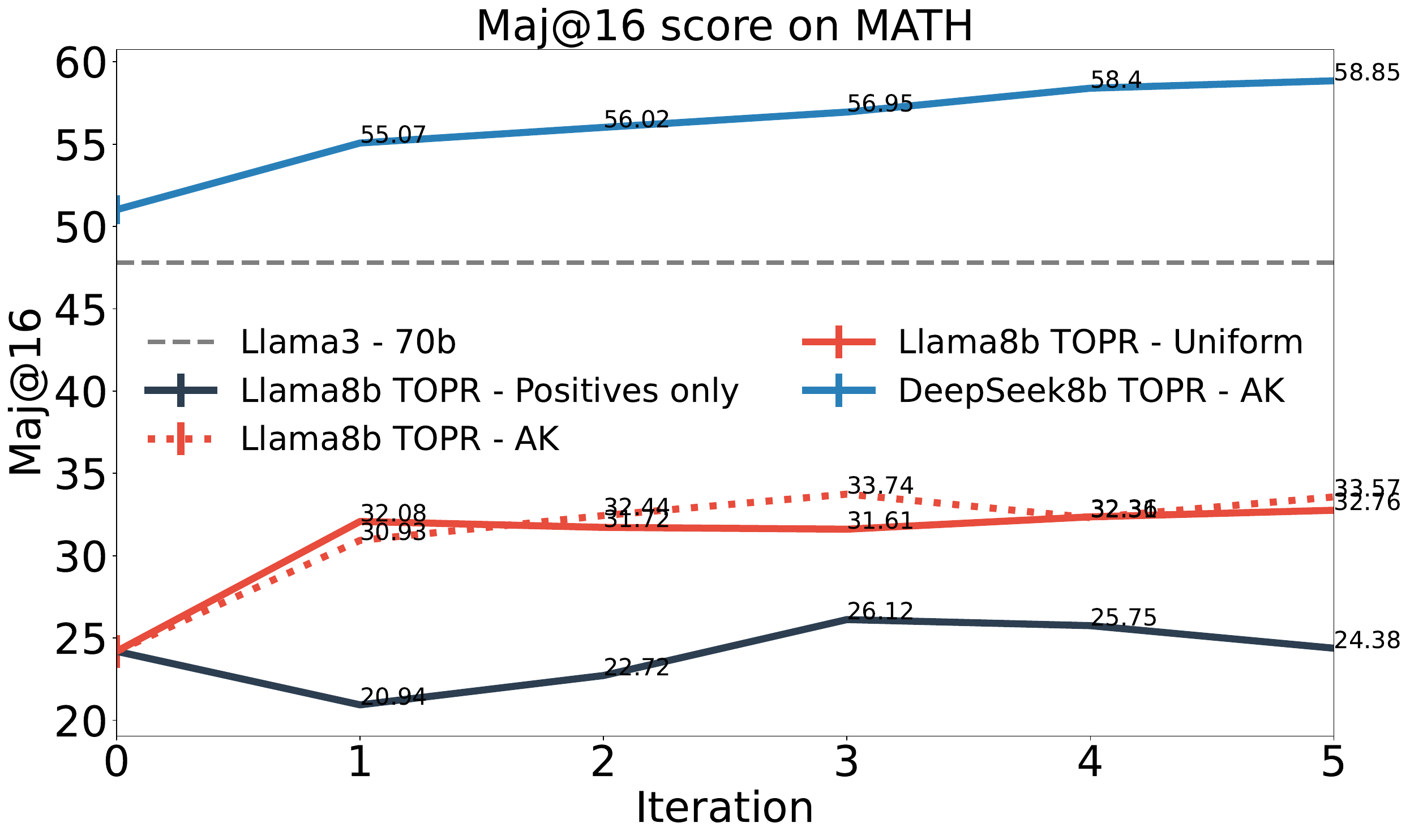}\caption{\textbf{Pass@1 (left) and maj@16 (right) scores on GSM8K (top) and MATH (bottom) for uniform sampling, positive-only sampling, and Anna Karenina sampling (see main text).} By combining TOPR and Anna Karenina sampling, we are able to fine-tune the DeepSeek-R1 8B model to achieve performance slightly superior to Llama 3 70B.
    \label{fig:gsm8k_multiiter}}
\end{figure}

\paragraph{TOPR outperforms across multiple iterations.} As a final experiment, we combine insights from our previous experiments to demonstrate that TOPR is an effective inner-loop algorithm for iterated fine-tuning of language models.
Training begins with a base model $\pi_0$, from which a dataset is sampled ($\mu_i = \pi_{i-1}, i = 1, 2, \dots$). We then subsample this dataset to create an iteration batch of to $N=50,000$ data points, and use these to train a new policy $\pi_i$ starting from $\pi_{i-1}$ as a reference. Figure~\ref{fig:gsm8k_multiiter} shows how model performance continues to improve over multiple iterations, both for GSM8K and MATH; furthermore, TOPR enables faster learning than positive-only sampling, allowing us to surpass DeepSeek 8B-level maj@16 accuracy within a few iterations.

One challenge with fine-tuning models that already perform quite well is that, as training progresses, the model is essentially presented with examples (e.g., math problems) that it already performs quite well on. This can make training quite inefficient. To combat this, we introduce a dataset-balancing technique we call \emph{Anna Karenina sampling}, based on Tolstoy's famous ``All happy families are alike; each unhappy family is unhappy in its own way.'' For each problem, we sample 64 candidate solutions, of which we only keep the \emph{first} positive example. The iteration batch is then filled with negative examples chosen at random from those candidates. On GSM8K, we find that this technique enables substantially more efficient learning (\textbf{79.6\%} pass@1 accuracy) compared to uniform sampling (\textbf{75.4\%}). The technique is less effective on early MATH iterations, where the model has low pass@1 accuracy and every positive example counts.
As further evidence of TOPR's effectiveness, applied to the more recent DeepSeek 8B model \citep{deepseek} it produces a model that rapidly surpasses the 70B version of Llama 3 in maj@16 accuracy.

\subsection{Learning to verify}

\citet{zhang2024generative} and \citet{mahan2024generative} recently studied the use of multiple CoT generations to verify the output of an LLM. In the context of math reasoning benchmarks, this \emph{generative verifier} acts as a reward model that is used in a best-of-n selection scheme, improving performance over self-consistency \citep{wang2022self}. Our next series of experiments aimed to study whether \emph{TOPR can improve verifier performance and improve solution quality for harder problems}. 

For each training sample in the MATH dataset, we used the 70B model to generate 16 solutions per problem, each with 4 verifications. We then fine-tuned an 8B model using TOPR to act as a generative verifier \citep{zhang2024generative} using a total of 480,000 data points. We evaluated this generative verifier in a weighted self-consistency setting, where four verifications are aggregated into a score for each solution, and the answer with the highest score sum is selected.
\cref{tab:MATH_model_comparison} shows that this procedure indeed produces a much more effective verifier for MATH generations, both in terms of its verifier accuracy and effect on solution quality. In an even more pronounced version of the results from Fig.~\ref{fig:gsm8k_G8b_with_bad_answers}, we find that TOPR fine-tunes the model to output almost no invalid generations -- simply because it is negatively rewarded for doing so.

\begin{table}
    \centering
    \begin{tabular}{rccc}
        \toprule
        Verifier & Verifier accuracy & Invalid rate & Weighted SC \\
        \midrule
        \textbf{None} & -- & -- & 55.5\% \\
        \textbf{Llama 3 8B} & 32.6\% & 34.2\% & 56.7\% \\
        \textbf{8B TOPR} & \textbf{70.9\%} & \textbf{0.90\%} & \textbf{61.5\%} \\
        \bottomrule
    \end{tabular}
    \caption{\textbf{Performance of a verifier trained with TOPR compared to using a base model verifier or no verifier.} Here ``Verifier accuracy'' is the pass@1 accuracy of the verifier (whether it correctly judges that a solution is right), ``Invalid rate'' is the number of verifications that could not be parsed successfully, and ``Weighted SC'' is weighted self-consistency with 32 generations (see e.g. \citep{luo2024improve}). When no verifier is used, Weighted SC indicates the usual maj@32 accuracy.}
    \label{tab:MATH_model_comparison}
\end{table}

\section{Conclusion and future work}
Our results show that a simple but principled change to REINFORCE is all that is needed to deploy it successfully and stably in the off-policy regime. Compared to existing dataset curation methods, our approach is more efficient: when generating the dataset as all data points are kept; at training time because no KL regularization is required, and negative examples are effectively made use of to improve performance; and at test time, because fewer solutions need to be generated. Our theory further provides an alternative, optimization-based perspective on truncated importance sampling for reinforcement learning, which may warrant revisiting other algorithms that make use of it \citep{munos16safe}. Finally, our analysis sheds new light on the role of the baseline parameter and dataset composition in off-policy reinforcement learning.

From here, there are a number of future avenues for research. On one hand, we limited our experimental work to the setting where $\mu$ is the model at the beginning of an iteration, and data points are generated in the ``self-taught'' style. It would naturally be beneficial to deploy this in the offline setting \citep{levine2020offline} with a different $\mu$, but at first glance this poses numerical challenges. We also limited ourselves to the training of large language models, but there is no reason to believe that TOPR would not perform equally in other application areas of reinforcement learning, from video games \citep{bellemare13arcade} to robotics \citep{kalashnikov2018scalable}.

\section{Author Contributions}
N.L.R. and M.G.B. developed the algorithm, provided the theoretical analysis, structured the experimental work, and led the overall project. N.L.R., M.G.B., J.L., A.B., and J.G. implemented the algorithm, the experimental framework, and performed experiments. N.L.R., M.G.B., and J.L. wrote the paper. J.L., J.G., A.F., C.P., E.T.L., S.W., and S.T. contributed to the technical infrastructure (datasets and software) with which the experimental work was performed.

\section*{Acknowledgments}
The authors would like to thank Matthieu Geist, Sal Candido, Rishabh Agarwal, Michael Bowling, Jesse Farebrother, Nathan Rahn, Harley Wiltzer, Charline Le Lan, and John Schulman for feedback and helpful discussions, as well as Matt Leus, Karl Moritz Hermann, and Nasib Naimi for support on this project.

N.L.R. did this work while working as a consultant for Reliant AI, using their compute infrastructure. This work was partially supported by a Canada CIFAR AI Chair.

\bibliography{arxiv_topr}
\bibliographystyle{plainnat}
\newpage
\appendix

In this appendix, we provide additional theoretical results, especially around the use of baselines in off-policy optimization.

\section{Proof of Proposition~\ref{prop:naive_reinforce_loss}}
As in the main text, we separate positive and negative trajectories as $T^+ := \{ \tau : R(\tau) \ge 0\}$ and $T^- := \{ \tau : R(\tau) < 
0\}$. Define the reward-weighted distribution
\begin{equation*}
    \murneg(\tau) = \left \{ \begin{array}{ll}
        \frac{\mu(\tau) |R(\tau)|}{\rmuneg} &         \text{if } R(\tau) < 0, \\
        0 & \text{otherwise;}
    \end{array} \right .
    \hspace{4em} 
        \rmuneg = \sum\limits_{\tau \in T^-} \mu(\tau) \big |R(\tau)\big| ,
\end{equation*}
and symmetrically for $\murpos$ and $\rmupos$.\footnote{If $\rmupos=0$ (resp. $\rmuneg=0$), our argument holds for any $\murpos$ (resp. $\murneg$).} We have
\begin{align*}
    J_{\mu,c}(\pi) - J(\mu) &= \expect_{\tau \sim \mu} \big [ R(\tau) - c \big ] \log \frac{\pi(\tau)}{\mu(\tau)} \\
    &= \expect_{\tau \sim \mu}  \left [ R(\tau) \log \frac{\pi(\tau)}{\mu(\tau)} \right ] - c \expect_{\tau \sim \mu} \left [ \log \frac{\pi(\tau)}{\mu(\tau)} \right ] \\
    &= \sum\nolimits_\tau \mu(\tau) R(\tau) \log \pi(\tau) + c \kl(\mu \cdbar \pi) + C_0 ,
\end{align*}
where $\kl(\mu \cdbar \pi)$ denotes the Kullback-Leibler divergence from $\mu$ to $\pi$ and $C_0 \in \bR$.
We now break the first term of the above equation into its positive and negative components:
\begin{align*}
\sum\nolimits_\tau \mu(\tau) R(\tau) \log \pi(\tau) &= \sum_{\tau \in T^+} \mu(\tau) R(\tau) \log \pi(\tau) + \sum_{\tau \in T^-} \mu(\tau) R(\tau) \log \pi(\tau) \\
&= \rmupos \sum_{\tau \in T^+} \murpos(\tau) \log \pi(\tau) - \rmuneg\sum_{\tau \in T^-} \murneg(\tau) \log \pi(\tau) \\
&= -\rmupos\kl(\murpos \cdbar \pi) + \rmuneg\kl(\murneg \cdbar \pi) + C_1 .
\end{align*}
Putting it all together, we see that
\begin{equation*}
    \mathcal{L}_{\mu,c}(\pi) = C + \rmupos \kl(\murpos \cdbar \pi) - \rmuneg \kl(\murneg \cdbar \pi) - c \kl(\mu \cdbar \pi) ,
\end{equation*}
with $C$ a constant independent of $\pi$.
\section{The impact of including negative examples}
\label{sec:appendix_negatives}
We recall the gradient of the TOPR objective:
\begin{align*}
    \nabla J_{\topr}(\pi) &= \sum_{\tau: R(\tau) > 0} \mu(\tau)\left[\frac{\pi(\tau)}{\mu(\tau)}\right]_{a^+}^{b^+} R(\tau)\nabla \log \pi(\tau) + \sum_{\tau: R(\tau) < 0} \mu(\tau)\left[\frac{\pi(\tau)}{\mu(\tau)}\right]_{a^-}^{b^-} R(\tau)\nabla \log \pi(\tau)
\end{align*}

Assume two trajectories: $\tau^+$ with positive return ($R(\tau^+)> 0$) and $\tau^-$ with negative return ($R(\tau^-)< 0$). Let us call $z^+$ and $z^-$ the associated logits, i.e. $\pi(\tau^+) = \frac{\exp(z^+)}{\sum_z \exp(z)}$, $\pi(\tau^-) = \frac{\exp(z^-)}{\sum_z \exp(z)}$.

\subsection{Exact gradient}
We first study the case of the exact gradient.
The total gradients for $z^+$ and $z^-$ are equal to:
\begin{align*}
    \nabla_{z^+}J_{\topr}(\pi) &= -\pi(\tau^+)\left(\sum_{\tau: R(\tau) > 0}\mu(\tau)\left[\frac{\pi(\tau)}{\mu(\tau)}\right]_{a^+}^{b^+}R(\tau) + \sum_{\tau: R(\tau) < 0}\mu(\tau)\left[\frac{\pi(\tau)}{\mu(\tau)}\right]_{a^-}^{b^-}R(\tau)\right) + \mu(\tau^+)\left[\frac{\pi(\tau^+)}{\mu(\tau^+)}\right]_{a^+}^{b^+}R(\tau^+)\\
    \nabla_{z^-}J_{\topr}(\pi) &= -\pi(\tau^-)\left(\sum_{\tau: R(\tau) > 0}\mu(\tau)\left[\frac{\pi(\tau)}{\mu(\tau)}\right]_{a^+}^{b^+}R(\tau) + \sum_{\tau: R(\tau) < 0}\mu(\tau)\left[\frac{\pi(\tau)}{\mu(\tau)}\right]_{a^-}^{b^-}R(\tau)\right) + \mu(\tau^-)\left[\frac{\pi(\tau^-)}{\mu(\tau^-)}\right]_{a^-}^{b^-}R(\tau^-)
\end{align*}

As $z^+$ and $z^-$ correspond to the logits, and are thus invariant to a shift, we study the difference between the gradients to understand the consequence of incorporating negative examples. We first call
\begin{align*}
    \Delta &= \left(\sum_{\tau: R(\tau) > 0}\mu(\tau)\left[\frac{\pi(\tau)}{\mu(\tau)}\right]_{a^+}^{b^+}R(\tau) + \sum_{\tau: R(\tau) < 0}\mu(\tau)\left[\frac{\pi(\tau)}{\mu(\tau)}\right]_{a^-}^{b^-}R(\tau)\right) \; .
\end{align*}
\begin{align*}
    \nabla_{z^+}J_{\topr}(\pi) - \nabla_{z^-}J_{\topr}(\pi) &= (\pi(\tau^-) - \pi(\tau^+))\Delta + \mu(\tau^+)\left[\frac{\pi(\tau^+)}{\mu(\tau^+)}\right]_{a^+}^{b^+}R(\tau^+) - \mu(\tau^-)\left[\frac{\pi(\tau^-)}{\mu(\tau^-)}\right]_{a^-}^{b^-}R(\tau^-) \; .
\end{align*}

SFT sets $a^-=b^-=0$. This has two impacts. First, $\Delta$ will be larger as the negative term in the sum is set to 0. Assuming that $\pi(\tau^-) - \pi(\tau^+) < 0$, this makes the first term in the difference of gradients more negative. Second, the term $- \mu(\tau^-)\left[\frac{\pi(\tau^-)}{\mu(\tau^-)}\right]_{a^-}^{b^-}R(\tau^-)$, which is positive for $b^- >0$, is set to 0. Overall, the difference of gradients is made smaller, leading to fewer differences between the gradient applied to positive trajectories and that applied to negative trajectories. Hence, less learning occurs, a phenomenon that can also be explained by baselines, see Sec.~\ref{sec:appendix_baselines}.

\subsection{Stochastic estimate of the gradient}
Another phenomenon occurs when only looking at a stochastic estimate of the gradient instead of the true expectation. After all, the goal of truncating the importance ratios is to deal with the variance induced by the stochasticity.

For this, we need to introduce two more trajectories. In addition to the two trajectories $\tau^+$ and $\tau^-$, which we will assume to be our two sampled trajectories, we consider $\nu^+$ and $\nu^-$, with associated logits $s^+$ and $s^-$, which have positive and negative returns (respectively) but have not been sampled. We have the following gradients:
\begin{align*}
    \nabla_{z^+}\hat{J}_{\topr}(\pi) &= \left[\frac{\pi(\tau^+)}{\mu(\tau^+)}\right]_{a^+}^{b^+}R(\tau^+)[1 - \pi(\tau^+)] - \left[\frac{\pi(\tau^-)}{\mu(\tau^-)}\right]_{a^-}^{b^-}R(\tau^-)\pi(\tau^+)\\
    \nabla_{z^-}\hat{J}_{\topr}(\pi) &= -\left[\frac{\pi(\tau^+)}{\mu(\tau^+)}\right]_{a^+}^{b^+}R(\tau^+)\pi(\tau^-) + \left[\frac{\pi(\tau^-)}{\mu(\tau^-)}\right]_{a^-}^{b^-}R(\tau^-)[1 - \pi(\tau^-)]\\
    \nabla_{s^+}\hat{J}_{\topr}(\pi) &= -\pi(\nu^+)\left(\left[\frac{\pi(\tau^+)}{\mu(\tau^+)}\right]_{a^+}^{b^+}R(\tau^+) + \left[\frac{\pi(\tau^-)}{\mu(\tau^-)}\right]_{a^-}^{b^-}R(\tau^-)\right)\\
    \nabla_{s^-}\hat{J}_{\topr}(\pi) &= -\pi(\nu^-)\left(\left[\frac{\pi(\tau^+)}{\mu(\tau^+)}\right]_{a^+}^{b^+}R(\tau^+) + \left[\frac{\pi(\tau^-)}{\mu(\tau^-)}\right]_{a^-}^{b^-}R(\tau^-)\right) \; .
\end{align*}
The last two equations correspond to the gradient with respect to trajectories that are not included in the sample. For each of these equations, the two terms inside the parentheses are of opposite signs. Hence, including both will generally decrease the norm of the gradient of unseen trajectories. As seen in the expected gradient, including both terms also increases the gradient of the positive trajectory $\tau^+$ while decreasing that of the negative trajectory $\tau^-$. In other words, setting $b^- > 0$ takes mass away from the gradients of unseen trajectories to gradients of seen trajectories. This will make the changes to the policy more localized.

\section{The impact of the baseline on KL regularization in TOPR}
\label{sec:appendix_baselines}
In standard REINFORCE, adding a baseline to the reward does not change the unbiasedness of the gradient estimate but affects its variance. The same is not true in off-policy policy optimization. 
Looking back at Eq.~\eqref{eq:topr_grad_general}, assuming $a^+=a^-=0$, $b^+=b^-=b$, we see that adding a baseline $c$ yields
\begin{align*}
\label{eq:topr_grad_general_baseline}
    \nabla J_{\topr}(\pi,c) &= \sum_{\tau} \mu(\tau)\left[\frac{\pi(\tau)}{\mu(\tau)}\right]_0^b [R(\tau) - c]\nabla \log \pi(\tau)\\
    &= \sum_{\tau} \mu(\tau)\left[\frac{\pi(\tau)}{\mu(\tau)}\right]_0^b R(\tau)\nabla \log \pi(\tau)\\
    &\qquad - \sum_{\tau} \mu(\tau)\left[\frac{\pi(\tau)}{\mu(\tau)}\right]_0^b c\nabla \log \pi(\tau)\\
    &= \nabla J_{\topr}(\pi,0)\\
    &\qquad - c\sum_{\tau} \mu(\tau)\frac{\pi(\tau)}{\mu(\tau)}\nabla \log \pi(\tau)\\
    &\qquad - c\sum_{\tau} \mu(\tau)\left(\left[\frac{\pi(\tau)}{\mu(\tau)}\right]_0^b - \frac{\pi(\tau)}{\mu(\tau)}\right)\nabla \log \pi(\tau)\\
    \nabla J_{\topr}(\pi,c) &= \nabla J_{\topr}(\pi,0) + c\sum_{\tau} \mu(\tau)\left(\frac{\pi(\tau)}{\mu(\tau)} - \left[\frac{\pi(\tau)}{\mu(\tau)}\right]_0^b\right)\nabla \log \pi(\tau)
\end{align*}

Assume a negative baseline $c$. Since $\frac{\pi(\tau)}{\mu(\tau)} - \left[\frac{\pi(\tau)}{\mu(\tau)}\right]_0^1$ is positive when $\pi(\tau) > b\mu(\tau)$, the additional term will \emph{decrease} $\pi(\tau)$ in that case. Hence, a negative baseline will discourage $\pi(\tau)$ to be above $b\mu(\tau)$ for all trajectories $\tau$. In that sense, it acts as a softer version of a KL regularizer that only alters $\pi$ when it deviates too much from $\mu$.
Alternatively, a positive baseline will encourage $\pi$ to be large, making the policy more deterministic. Note that this effect is solely due to clipping and goes against the effect due to stochasticity observed by~\citet{chung2021beyond}.

\section{From reinforcement learning to supervised learning}
\label{sec:appendix_rl_supervised}
While it was presented as a standard RL objective, Eq.~\ref{eq:rl_supervised_loss} actually meshes a supervised learning part, the sum over examples $x_j$, and a reinforcement learning part, the expectation over prompt completions $y$ given an example $x_j$. While we discussed the policy from which the $y$'s were sampled, we have not discussed which distribution over $x$ to use. Instead of changing the weight of each training sample, as can be done in supervised learning, the RL framework offers the possibility of sampling a different number of trajectories for each question.

\citet{duchi2019variance} proposed a generalization bound depending on the empirical variance of the loss across training examples: a more uniform loss across training samples leads to a better generalization error. In our scenario, each question $x$ is associated with a different difficulty: the proportion of correct completions $y$ generated by the base model. It is thus natural to generate \emph{more} completions for the hard questions, those with few correct completions, than for the easier ones, as that would encourage the model to spend more capacity on these harder examples, reducing the gap in loss between all examples. Keeping only the correct completions, as done by STaR and other popular methods, does the exact opposite: the training set is mainly comprised of completions from the \emph{easy} questions as they have more positive ones. This could widen the performance gap among examples.

\begin{figure}
    \centering
    \includegraphics[width=0.9\linewidth]{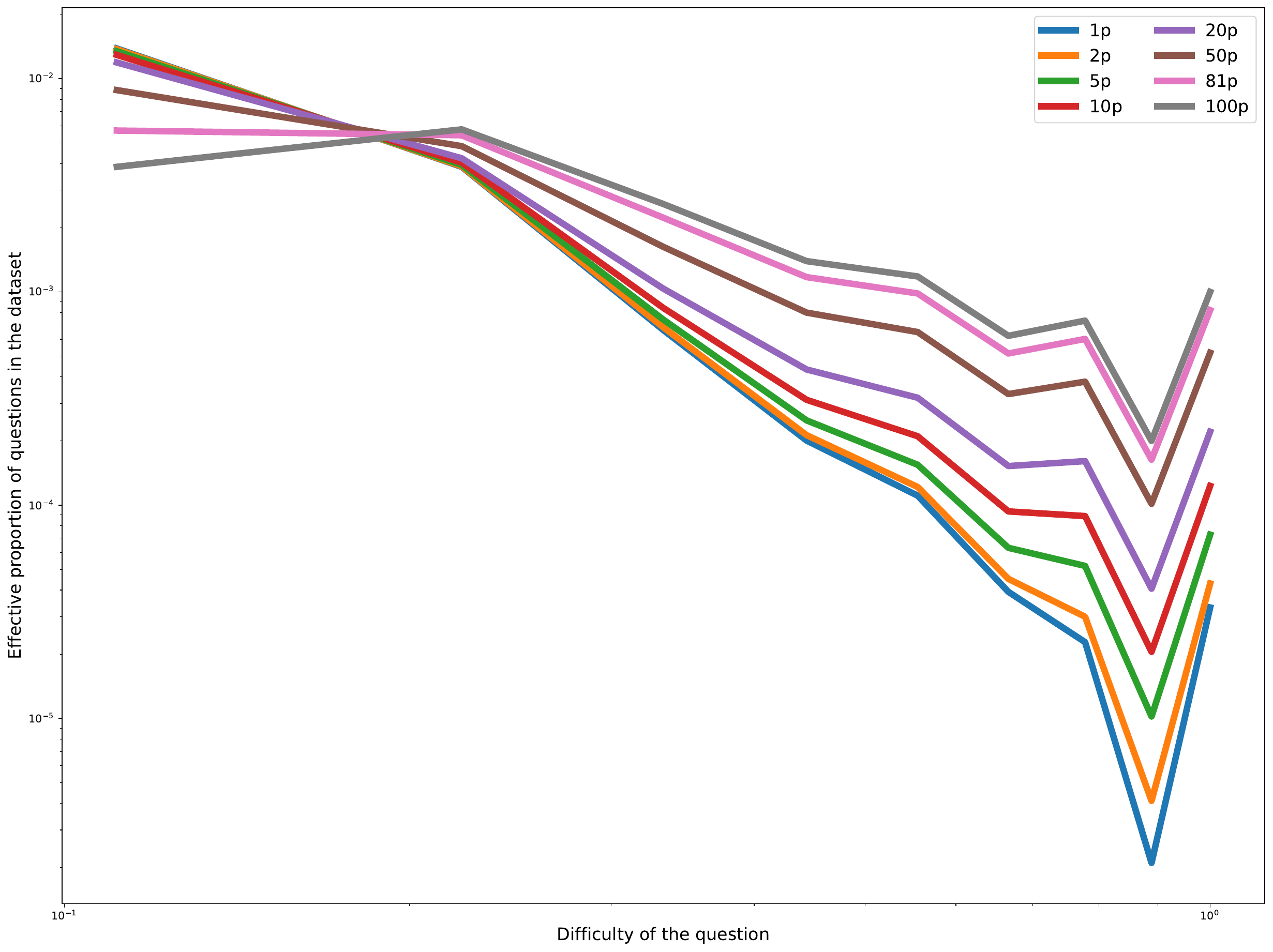}
    \caption{Effective proportion of questions of each difficulty in the dataset for various effective proportions of positive examples. Datasets with fewer effective positive examples give more weight to difficult questions.}
    \label{fig:dataset_difficulty}
\end{figure}

Fig.~\ref{fig:dataset_difficulty} shows the importance given to easy and hard questions when varying the baseline, i.e. the effective proportion of positive examples. For each question in the training set, we generated 128 solutions, then created 10 equally-spaced buckets based on the expected accuracy of the base model for these questions. We then computed, on average, how many solutions each of the question in a specific bucket had. To account for the baseline parameter, we gave a weight of $(1-c)$ to each correct solution and a weight of $(1+c)$ to each incorrect solution.

We see that, as $c$ increases, i.e. the effective proportion of positive examples decreases, more (relative weight) is given to difficult questions. There seems to be a correlation between the performance of TOPR and the shape of curve in Fig.~\ref{fig:dataset_difficulty}. 

\end{document}